\documentclass{article} 
\usepackage{iclr2026_conference,times}
\usepackage{algorithm}
\usepackage{algorithmic}
\usepackage{subfigure}
\usepackage{soul}
\usepackage{color}
\usepackage{xcolor}
\usepackage{amsmath}
\usepackage{amsfonts}
\usepackage{multirow}
\usepackage{wrapfig}
\usepackage{graphicx}
\usepackage{xspace}
\usepackage{booktabs}
\usepackage{enumitem}
\usepackage{nameref}
\usepackage{array}
\usepackage{nicematrix} 
\usepackage{booktabs}
\usepackage{hyperref}
\usepackage[capitalize,noabbrev]{cleveref}
\usepackage[flushleft]{threeparttable}
\usepackage{multirow}
\usepackage{enumitem}
\usepackage{siunitx}
\usepackage{caption}
\usepackage{wrapfig}
\newcommand{\method}{\texttt{ODEBRAIN}\xspace }
\usepackage{caption}
\usepackage{booktabs,threeparttable}
\usepackage{siunitx}
\newcommand{\best}[1]{\mathbf{#1}}
\newcommand{\second}[1]{\underline{#1}}
\usepackage{newfloat}
\usepackage{listings}

\usepackage{amsmath,amsfonts,bm}

\def\eqref#1{equation~\ref{#1}}

\def\1{\bm{1}}

\def\vz{{\bm{z}}}

\DeclareMathAlphabet{\mathsfit}{\encodingdefault}{\sfdefault}{m}{sl}
\SetMathAlphabet{\mathsfit}{bold}{\encodingdefault}{\sfdefault}{bx}{n}

\usepackage{hyperref}
\usepackage{url}

\hypersetup{
    colorlinks,
    linkcolor={red!50!black},
    citecolor={brown!85!red},
    urlcolor={blue!80!black}
}

\usepackage[capitalize,noabbrev]{cleveref}

\title{ODEBrain: Continuous-Time EEG Graph for Modeling Dynamic Brain Networks}

\author{
Haohui Jia\textsuperscript{$\spadesuit$},
\hspace{0.5mm}Zheng Chen\textsuperscript{$\clubsuit $,$\dagger$},
\hspace{0.5mm}Lingwei Zhu\textsuperscript{$\diamondsuit$},
\hspace{0.5mm}Rikuto Kotoge\textsuperscript{$\clubsuit$},
\hspace{0.5mm}Jathurshan Pradeepkumar\textsuperscript{$\heartsuit$},
\\
\hspace{0.5mm}\textbf{Yasuko Matsubara}\textsuperscript{$\clubsuit $},
\hspace{0.5mm}\textbf{Jimeng Sun}\textsuperscript{$\heartsuit$},
\textbf{Yasushi Sakurai\textsuperscript{$\clubsuit $},
\hspace{0.5mm}Takashi Matsubara\textsuperscript{$\spadesuit$}}
\\
\hspace{0.5mm}\textsuperscript{$\spadesuit$}Information Science and Techinology, Hokkaido University, Japan\\
\hspace{0.5mm}\textsuperscript{$\clubsuit $}SANKEN, The University of Osaka, Japan\\
\hspace{0.5mm}\textsuperscript{$\diamondsuit$}Great Bay University, China\\
\hspace{0.5mm}\textsuperscript{$\heartsuit$}Department of Computer Science, University of Illinois Urbana-Champaign, USA\\
\hspace{0.5mm}\textsuperscript{$\dagger$}Corresponding author: \texttt{chenz@sanken.osaka-u.ac.jp}
}

\iclrfinalcopy 
\begin{document}

\maketitle

\begin{abstract}
Modeling neural population dynamics is crucial for foundational neuroscientific research and various clinical applications. Conventional latent variable methods typically model continuous brain dynamics through discretizing time with recurrent architecture, which necessarily results in compounded cumulative prediction errors and failure of capturing instantaneous, nonlinear characteristics of EEGs. We propose \method, a Neural ODE latent dynamic forecasting framework to overcome these challenges by integrating spatio-temporal-frequency features into spectral graph nodes, 
followed by a Neural ODE modeling the continuous latent dynamics.
Our design ensures that latent representations can capture stochastic variations of complex brain states at any given time point.
Extensive experiments verify that \method can improve significantly over existing methods in forecasting EEG dynamics with enhanced robustness and generalization capabilities.

\end{abstract}

\section{Introduction}
\label{sec:introduction}

Modeling dynamic activity in brain networks or connectivity using electroencephalograms (EEGs) is crucial for biomarker discovery~\citep{brain_dynamics, Jones2022computational} and supports a wide range of clinical applications~\citep{kotoge2024splitsee,pradeepkumar2025TFM}.
Temporal graph networks (TGNs), which integrate temporally sequential models (such as RNNs) with graph neural networks (GNNs), have recently emerged as a promising approaches \citep{GNN_ICLR22,GNN_AAAI23,delavari2024synapsnet,li2024amag}.
These methods represent multi-channel EEGs as graphs, where GNNs capture spatial dependencies and sequential models capture fine-grained temporal dynamics, thereby providing mechanistic insights into gradual evolution of brain networks.

\begin{figure}[t]
    \centering\includegraphics[width=0.99\linewidth]{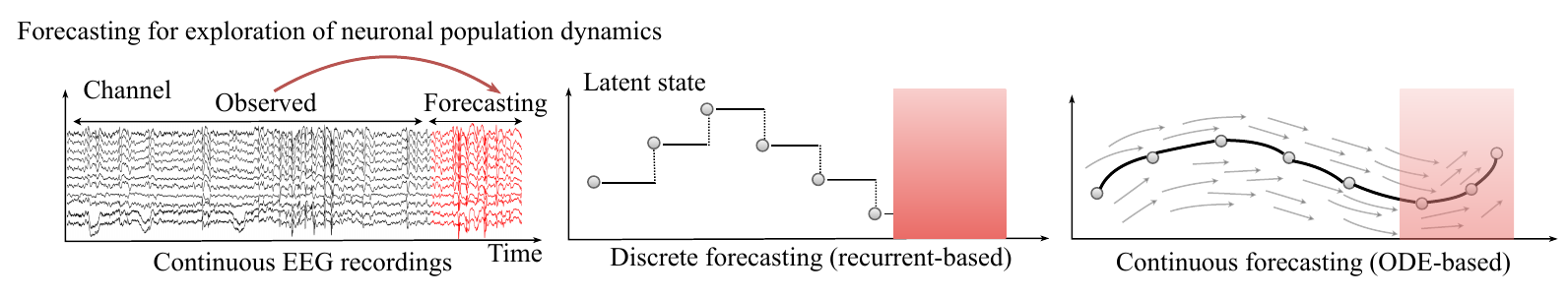}
   \caption{(Left) Continuous EEG real-time neuronal activity recordings. (Mid) Recurrent-based methods employ discrete modeling.
   (Right) ODE provides a continuous representation for forecasting neuronal population dynamics.}
   \label{fig:figure1}
\end{figure}

However, a critical yet overlooked problem remains: existing methods transform EEGs into fixed, discrete time steps, which conflicts with the inherently continuous nature of dynamic brain networks.
Such discretization imposes rigid windowing assumptions and prevents models from capturing the unfolding time-course dynamics or irregular transitions in brain networks, as in Figure  \ref{fig:figure1}.
This paper aims to tackle this issue by developing a novel method that models EEGs in an explicitly continuous manner, while  leveraging Neural Ordinary Differential Equations (NODEs) \citep{chen2018neural}.

In contrast to RNN-based sequential models, which discretize time into fixed steps, NODEs parameterize the derivative of the hidden state and integrate it continuously over time~\citep{park2021vid}.
This formulation provides a principled method to model the dynamical evolution of neural activity~\citep{hu2024modeling} and has been studied across domains~\citep{hwang2021climate}, including brain imaging~\citep{han2024brainode}.
In this paper, we study a novel and critical problem: modeling dynamics brain networks with NODEs to learn informative continuous-time representations from EEGs.
This remains a unexplored and non-trivial task, and we focus on two main challenges:\\

$\rm(\hspace{.18em}i\hspace{.18em})$ \textit{Effective spatiotemporal modeling for ODE initialization.}
NODEs critically depend on the quality of their initial conditions, because the ODE solver propagates trajectories starting from this initialization. Accordingly, a poor initialization propagates errors and destabilizes long-term dynamics. 
However, EEG signals are inherently noisy and stochastic, which makes learning robust spatiotemporal representations of brain networks particularly challenging. Therefore, designing an initialization that captures meaningful spatiotemporal structures is essential for stable ODE integration and effective downstream learning.
\\
$\rm(\hspace{.18em}ii\hspace{.18em})$ \textit{Accurate trajectory modeling.}
Trajectory modeling is essential for NODEs, as their strength lies in learning continuous latent dynamics rather than discrete predictions.
Unlike conventional time-series data, which often exhibit stable patterns such as periodicity or long-term trends \citep{klotergens2025physiomeode}, EEG signals are highly variable, which makes trajectory learning particularly challenging. Thus, a major challenge is to constrain and preserve meaningful trajectories in the latent space so that NODEs can accurately capture the continuous dynamics of EEGs.

In this paper, we introduce a new continuous-time EEG Graph method, \method, based on the NODE, for modeling dynamic brain networks.
To address the above challenges, we \textbf{first} propose a dual-encoder architecture to provide effective initialization for NODEs.
One encoder captures deterministic frequency-domain observations to model brain networks, whereas the other integrates raw EEG representations to retain stochastic characteristics. 
This combination yields robust spatiotemporal features for initializing the ODE solver.
\textbf{Second}, we introduce a trajectory forecasting decoder that maps latent representations obtained from NODE solutions back to graph structures. Thereafter, a multistep forecasting loss function was applied to explicitly predict future brain networks at different time steps. This design enables direct trajectory modeling of dynamic brain networks and enhances accuracy.
\textbf{Third}, beyond modeling, we are the first to propose the use of the gradient field of NODEs as a metric to quantify EEG brain network dynamics. 
Herein, we also conduct a case study on seizure data to illustrate the clinical interpretability of this method.

\begin{itemize}[left=0pt]
    \item \textbf{New problem Formulation.}
    To the best of our knowledge, we are the first to explicitly formulate EEG brain networks as a continuous-time dynamical system, where the brain network is represented as a sequence of time-varying graphs whose latent dynamics are governed by a NODE.
    This perspective differs from prior approaches based on recurrent models, which model gradual state transitions in a discrete-time framework rather than in a principled continuous-time manner.
    \item \textbf{Novel Method.}
    We develop the \method framework that integrates three key components.
    It first combines deterministic graph-based features with stochastic EEG representations to produce a robust initial state.
    Then an explicit trajectory forecasting decoder with multi-step forecasting loss hat models temporal–spatial dynamics continuously, enabling principled forecasting of evolving brain networks.
    \item \textbf{Comprehensive Evaluation.} 
    We demonstrate strong performance across benchmarks and provide retrospective clinical case studies highlighting the interpretability.
    Our \method outperforms all baselines on the TUSZ dataset, achieving 6.0\% and 8.1\% improvements in F1 and ACC, respectively. 
    On the TUAB, \method consistently achieves best performance, such as 1.2\% improved F1 and 2.4\% improved AUROC. Moreover, we further evaluate the learned field and its clustering to reveal the dynamic behaviors (varying speed and direction) between seizure and normal states, and achieving 12.0\% improvement for brain connectivity prediction.
\end{itemize}

\section{Related Works}
\label{sec:related works}

\noindent\textbf{Temporal graph methods for modeling EEG dynamics.}

GNNs have emerged as powerful method for effectively capturing spatial dependencies and relational structures in the analysis of brain networks \citep{10020662,yang2022unsupervised,kan2023r}. 
Specifically, EEG-GNN performs a learnable mask to filter the graph structure of EEG for cognitive classification tasks \citep{9630194}.  
ST-GCN formulates the connectivity of spatio-temporal graphs to capture non-stationary changes \citep{gadgil2020spatio}. 
\citet{GNN_ICLR22} have introduced the DCRNN approach for graph modeling, setting a new standard for SOTA in seizure detection and classification tasks. 
Following this, GRAPHS4MER \citep{pmlr-v209-tang23a} enhanced the graph structure and integrated it with the MAMBA framework to improve long-term modeling capabilities. 
AMAG \citep{li2024amag} forecasting method has been proposed to effectively capture the causal relationship between past and future neural activities, demonstrating greater efficiency in modeling dynamics.
More recently, EvoBrain investigates the expressive power of TGNs in integrating temporal and graph-based representations for modeling brain dynamics \citep{kotoge2025dynamic}.
However, these studies rely on discrete modeling, and may lead to suboptimal representation of continuous dynamics of brain networks.

\noindent\textbf{Differential equations for brain modeling. }
Modeling brain function as low-dimensional dynamical systems via differential equations has been a long-standing direction in neuroscience~\citep{churchland2012neural,mante2013context,vyas2020computation}, and nonlinear EEG analysis for brain activity mining \citep{pijn1997nonlinear, xue2016minimum,lehnertz2003seizure,lehnertz2008epilepsy,mercier2024value}. 
Recently, Neural ODEs (NODEs) formulate dynamical systems by parameterizing derivatives with neural networks and have shown impressive achievements across diverse fields \citep{fang2021spatial,hwang2021climate,park2021vid}.
In BCI and epilepsy modeling, controllable formulations and fractional dynamics provide important theoretical foundations for modeling brain dynamics \citep{gupta2018re,tzoumas2018selecting,lu2021detection,martis2015epileptic,lepeu2024critical}. In latent-variable dynamics models, the EEG and neuronal processes are described as fractional dynamics \citep{gupta2019learning,gupta2018dealing,yang2019data,yang2025spiking}.
In neuroscience, \citet{kim2021inferring} learn neural activities by modeling the latent evolution of nonlinear single-trial dynamics with Gaussian processes from neural spiking data. 
\citet{hu2024modeling} propose using a smooth 2D Gaussian kernel to represent spikes as latent variables and describe the path dynamics with linear SDEs. Another study~\citep{10230734} demonstrates robust performance in neuroimaging by combining biophysical priors with NODEs, starting from predefined cognitive states. \citep{chen2024eeg} have shown the advantage of graph ODE by modeling continuous-time propagation for EEG emotion task. \citet{han2024brainode} further illustrate that integrating spatial structure with NODEs can effectively facilitate the modeling of neuroimaging dynamics, even in the presence of missing data.
However, these studies focus on imaging data or neuronal feature engineering, while data-driven modeling of brain networks with fractional dynamics from EEGs remains underexplored.


\section{Preliminary and Problem Formulation}
\label{sec: problem}

\textbf{Neural Ordinary Differential Equations.}
NODEs~\citep{chen2018neural} provide a framework for modeling continuous-time dynamics by parameterizing the derivative of a hidden state using neural networks.  
Intuitively, NODEs solve the trajectory of the hidden state continuously at any arbitrary time $\tau$, rather than restricting updates to fixed discrete steps $\Delta t$ in RNNs.  
Specifically, the hidden dynamics are computed via an adaptive numerical ODE solver:
\begin{align}
\boldsymbol{z}(t+1) &\simeq \texttt{ODEsolver}(\boldsymbol{z}_0,f_{\theta}) 
= \boldsymbol{z}_0 + \int_{t}^{t+1} f_{\theta}(t,\boldsymbol{z}_t)\, dt ,
\label{ode func}
\end{align}
where $f_{\theta}$ is a continuous, differentiable function parameterized by a neural network.  
This formulation yields a unique continuous trajectory $\boldsymbol{z}(t)$ over an interval $[t_0, t_0+\tau]$.

\textbf{Intuition in Modeling EEG Dynamics.}
Conventional sequential models, such as RNNs, have been a standard tool for modeling EEG.
However, they implicitly assume that time can be discretized into fixed steps and that state transitions, such as the onset of a seizure, must occur exactly at these steps \citep{kotoge2025dynamic}.  
While this assumption simplifies the computation, it poorly matches the reality of EEG, where brain activity evolves continuously and transitions can occur at arbitrary points in time.  
In contrast, NODEs address this limitation by modeling EEG dynamics through a continuous function $f_\theta$ whose integration yields smooth latent trajectories.  
Within this framework, discrete EEG signals recorded at sampling intervals are treated as observations sampled from an underlying continuous process $\int f_\theta(t)\,dt$.  
This perspective allows NODEs to capture both gradual oscillatory rhythms and abrupt transitions in neural activity, thereby providing a more faithful representation of the EEG brain dynamics. 

However, applying the NODE to EEG is non-trivial, and we recognize two questions needing to be answered:

\begin{enumerate}[left=0pt]
    \item \textit{Robust initialization $\vz_0$ against transients and stochasticity in EEGs.}
    NODE requires a well-cablibrated starting condition $\vz_0$ to effectively forecast future behavior. This is because EEGs are highly stochastic, or even chaotic to an extent. Their key features are transient and may appear without any preindicator \citep{chenIJCAI}. Without proper initialization $\vz_0$ as a guide, the integration of the model $f_{\theta}$ over time alone cannot accurately forecast future states.
    \item \textit{Meaningful objectives of $f_{\theta}(t,\boldsymbol{z}_t)$ to capture the underlying EEG dynamics.}
    Standard NODE training typically relies on regression-like objectives aimed at forecasting future states.   
    A key challenge lies in identifying which representations best capture the underlying neural dynamics, so that $f_{\theta}(t,\boldsymbol{z}_t)$ is guided toward modeling the true evolution of brain networks rather than only surface-level predictions.
    For example, in seizure analysis, the model must also learn to discern not only seizure but also any leading states that herald an impending seizure \citep{NatNeuroscience2021}.
\end{enumerate}

\noindent\textbf{Problem Statement (Modeling Dynamic Brain Networks). }  
Given the observed EEG up to time $t$, denoted as $\mathbf{X}_{\leq t}$, the objective is to model the dynamics of the brain network and forecast their future evolution.  
The predicted dynamics act as representations of brain states, enabling the distinction between conditions such as seizure and non-seizure.  
Following prior work \citep{GNN_ICLR22,SODorAAAI2025}, we represent the brain as a graph and aim to develop an EEG-based NODE ($\Omega$) to predict a sequence of time-varying graphs:
\begin{equation}
\mathcal{G}_{t+1:t+K} = \{\mathbf{X}_{t+1}, \dots, \mathbf{X}_{t+K}\} 
= \Omega \bigl(\boldsymbol{z}_0,f_{\theta(\mathcal{G}_{1:t})} \bigr).
\end{equation}
Over the next $K$ steps graph, where $\mathcal{G}_{1: t}$ denotes the observed brain networks up to time $t$, and  
$\mathcal{G}_{t+1:t+K}$ represents the predicted dynamic brain networks.
These graphs characterize dynamic brain networks, but this problem poses two key challenges:  
(i) obtaining a robust initialization $\mathbf{z}_0$ that can resist the transient and stochastic nature of EEGs; and  
(ii) defining an objective for $f_\theta$ that faithfully captures the underlying neural dynamics.

\begin{figure*}[t]
\centering
\includegraphics[width = 0.95\textwidth]{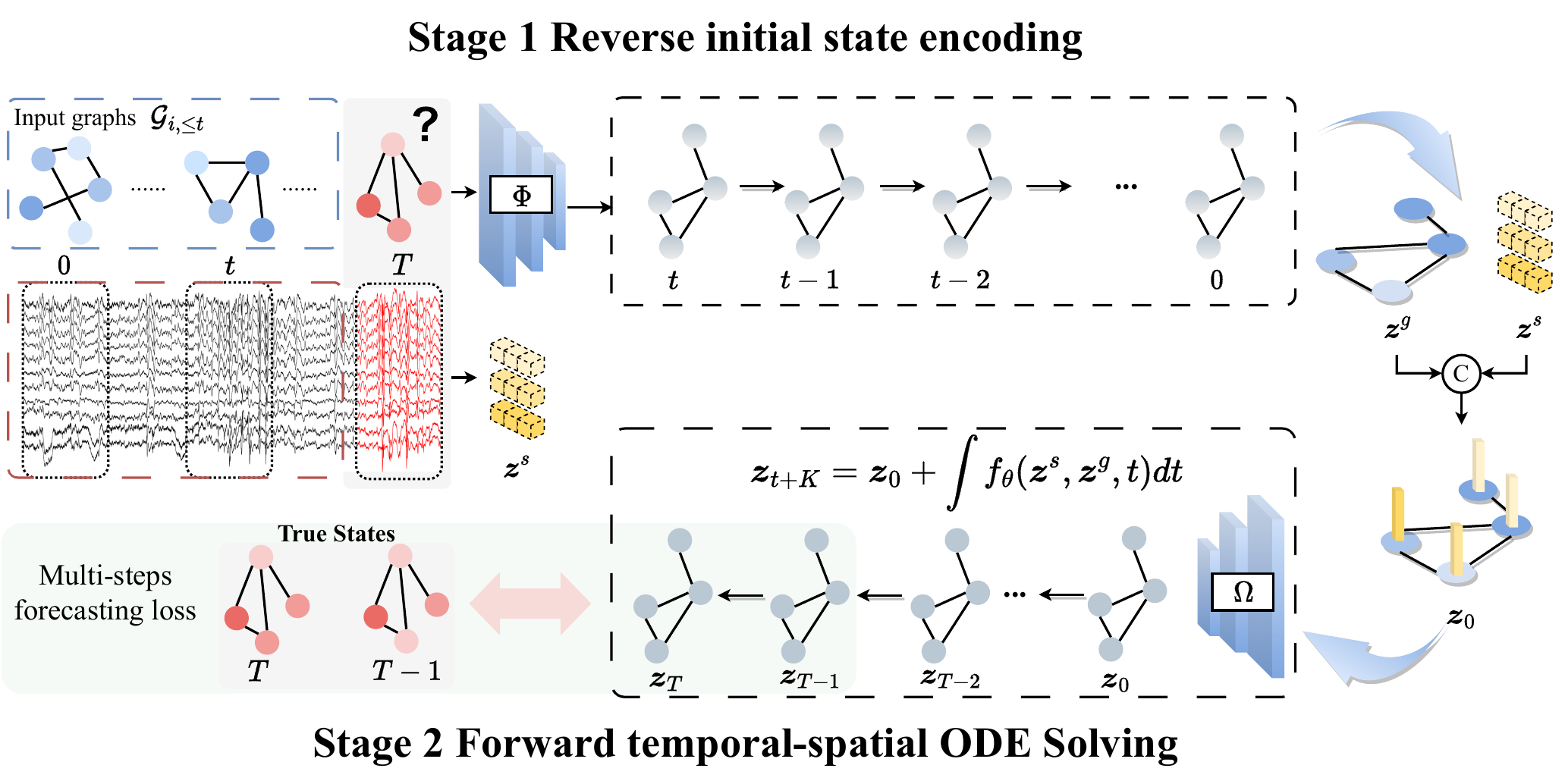}
\caption{Continuous neural dynamics modeling via \method with graph forecasting. In stage 1, multi-channel EEG signals are encoded into spectral graph snapshots and fused with raw features to construct noise-robust initial states for ODE integration, predicting future spectral graphs. In stage 2, \method propagates latent states through time, generating dynamic field $f$ that captures continuous trajectories. Finally, future graph node embeddings are obtained by $z_T$ and compared with the ground-truth graph nodes.}
\label{fig: proposal}
\end{figure*}
\label{sec: method}
\section{Methodology}
Figure \ref{fig: proposal} shows the system overview of \method.
Specifically, graph representations are obtained from each EEG segment (Section \ref{sec:graph}), entering stage 1: attaining reverse initial state encoding $\vz^g$ and temporal encoding $\vz^s$ (Section \ref{subsec:temporalspatialencoder}).
Stage 2 consists of a Neural ODE that takes as input $\vz^g, \vz^s$ (Section \ref{subsec:ODE}).
Finally, forecasting loss between ODE output and ground truth is computed.

\subsection{Stage 1: Reverse Initial State Encoding}
\label{subsec:temporalspatialencoder}

\textbf{Spectral Node Embedding. }
Previous discrete forecasting studies have shown that the capacity to estimate future neural dynamics depends on past activity in \citep{li2024amag}. We also define this forecasting paradigm within our \method. 
Intuitively, both the latent initial state $\boldsymbol{z}_{0}$ and the field $f$, i.e., $\frac{dz(t)}{dt}$ are described by encoding the past observation $\mathcal{G}_{i,\leq t}$ to govern the latent continuous evolution. 
The works of \citep{rubanova2019latent,chen2018neural} suggest that the construction of an effective latent initial state requires an autoregressive model capable of jointly extracting both the initial condition and the latent evolution.  
Accordingly, we introduce a graph state descriptor $\mathrm{\Phi}:\mathbb{R}^{d}\mapsto \mathbb{R}^{m}$ that represents latent graph states $\boldsymbol{z}^{g} \in \mathbb{R}^{m}$ using the autoregressive and graph network modules. 

Specifically, given the observations until now $\mathcal{G}_{i,\leq t}$ as input, we perform a sequence representation for the node and edge attributes.
For node embeddings, node evolution is computed by $\boldsymbol{h}_{i}^{n} = \text{GRU}^\text{node}(\mathcal{X}_{i,\leq t})$ where $\mathcal{V}_{i,\leq t}$ denote the spectral attribute sequences of node $i$ and $ \mathcal{X}_{i,\leq t}$ the spectral intensity.
Similarly, for edge, the attribute sequences are defined from adjacency matrices by
$\boldsymbol{h}_{ij}^{e}=\text{GRU}^\text{edge}(\mathcal{A}_{ij,\leq t})$.
The resulting node and edge embeddings are integrated into an aggregated graph structure ${\mathcal{G}} = (\boldsymbol{h}^{n}_{i,t},\boldsymbol{h}_{ij,t}^{e})$ to be learned by a GNN to capture the spatial dependency across epochs:
$\boldsymbol{z}^{g}=\text{GNN}\left(\boldsymbol{h}_{i}^{n},\boldsymbol{h}_{ij}^{e}\right)$.
The forward process of $\mathrm{\Phi}$ captures both the epoch variations between frequency bands and explicit channel correlations.

\noindent\textbf{Temporal Embedding with Stochasticity. } 
Accurate modeling of the temporal evolution of EEG signals is crucial because neural dynamics inherently exhibits nonuniform temporal fluctuations and asynchronous activations across channels. Although the graph descriptor $\mathrm{\Phi}$ effectively captures the evolution of the node and edge attributes, STFT segments the EEG signals by constant windows, which inevitably disrupts the continuous temporal correlation between the raw EEG observations.

\captionsetup{skip=2pt}
\begin{wrapfigure}[16]{r}{0.45\textwidth}
\centering
\includegraphics[width=0.96\linewidth]{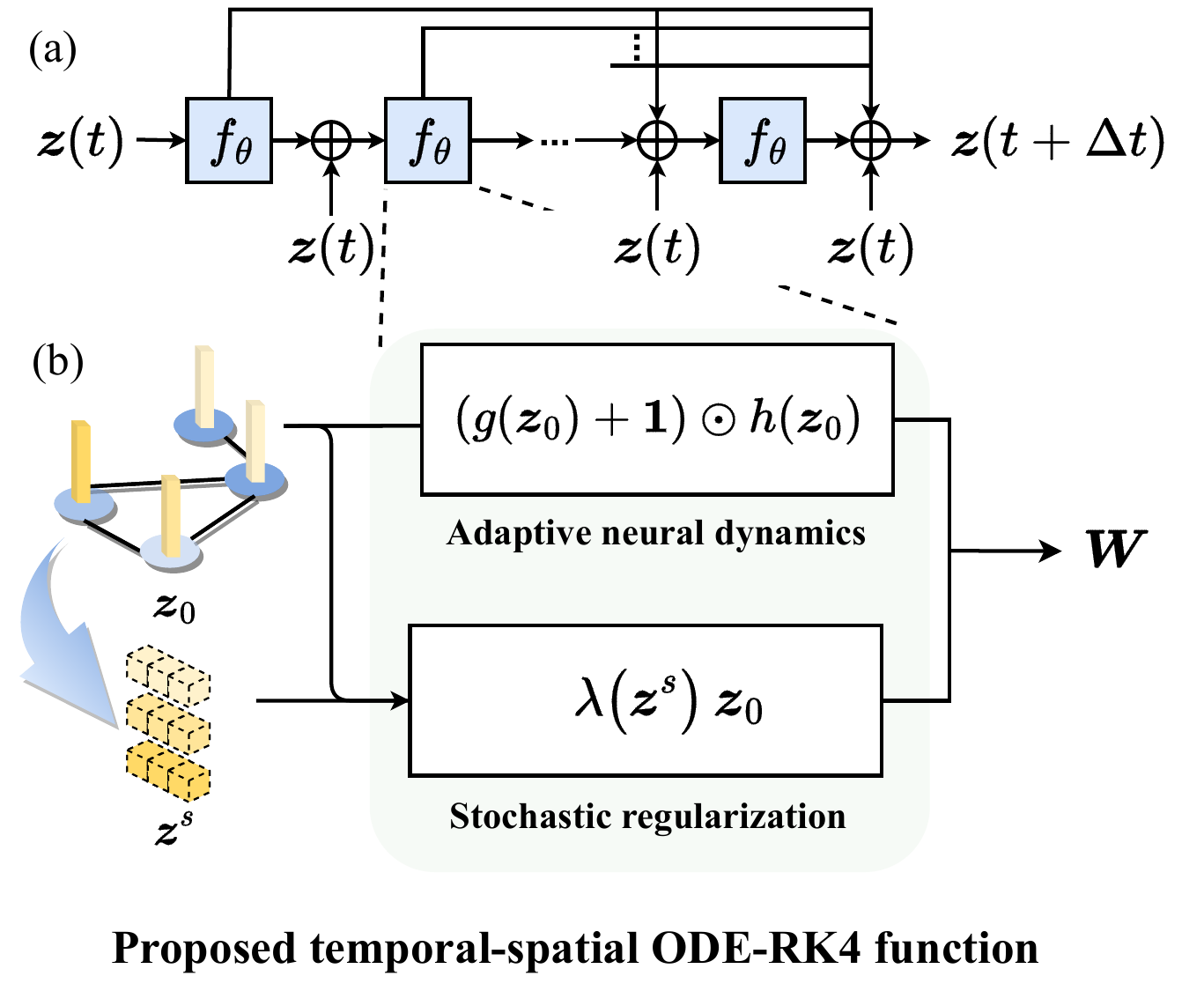}
\caption{The full structure of the temporal-spatial ODE solving. (a)RK-4 step numerical solver. (b) Procedure of temporal-spatial $f_{\theta}$.}
\label{fig:ST_ODE}
\end{wrapfigure}
Moreover, fully deterministic latent representations lack the flexibility necessary to effectively represent \textit{transient motions} of EEG as analyzed in Section \ref{sec: problem}. Conversely, introducing controlled randomness into temporal embeddings serves as a natural regularization strategy, effectively increasing the robustness and preventing premature convergence to suboptimal.
Here, we apply the temporal descriptor $\mathrm{\Psi}: \mathbb{R}^{T \times L} \mapsto \mathbb{R}^{c}, c \ll m$ to quantify the randomness of the raw EEG epochs across $N$ channels into $\boldsymbol{z}^{s} \in \mathbb{R}^{c}$. Given EEG segments $\mathbf{X}$ from $N$ channels within a sliding window length $L$, we define stochastic temporal embedding as $\boldsymbol{z}^{s} = \mathrm{\Psi}(\mathbf{X}_{T \times L,\leq N})$.
The controlled stochasticity further acts as a form of latent space regularization, enhancing generalization and robustness to noise in EEG data collection.

\subsection{Stage 2: Forward Temporal-Spatial ODE Solving}
\label{subsec:ODE}
Depending on the above encoding process, we define the initial state $z_{0} = [z^{s},z^{g}]$ with $\mathrm{\Phi}\circ\mathrm{\Psi} \mapsto \mathbb{R}^{m+c}$, which summarizes stochastic temporal variability and deterministic spectral connectivity, respectively.
Given the initial state $\boldsymbol{z}_{0} \in \mathbb{R}^{m+c}$, the general approach models the ODE vector field following the classical neural network solution $f_{\theta}$ with residual connection as:
\begin{equation}
    d\boldsymbol{z}(t) \equiv f_{\theta}(\boldsymbol{z}(t),t;\mathbf{\Theta})dt,\quad \boldsymbol{z}_0 = \left[ \boldsymbol{z}^{s},\boldsymbol{z}^{g} \right], \quad t \in [t+1, t+K] \label{our: IP}
\end{equation}
where $f_{\theta}: \mathbb{R}^{m+c} \mapsto \mathbb{R}^{m+c}$ represents a vector field to capture complicated dynamics and its continuous evolution is governed by $f_{\theta}$ with the learnable $\mathbf{\Theta}$ across the entire epoch sequences. However, this introduces the challenge of optimizing the deep network-based $f_{\theta}$ over highly variable EEG states, making large solver errors.

Considering the deep architecture-based multi-step numerical solver design \citep{lu2018beyond, oh2024stable} and logic gating interaction of brain dynamics \citep{goldental2014computational}, we design a temporal-spatial ODE solution to incorporate the initial state $\boldsymbol{z}_{0}$ for additive and gate operations as shown in Figure \ref{fig:ST_ODE}. In addition, we further introduce an adaptive decay component conditioned on the stochastic temporal state $\boldsymbol{z}^{s}$, to adjust the vector field $f_{\theta}$, accounting for the complexity and dynamic nature of the brain as a system.
As shown in Figure \ref{fig:ST_ODE}(b), the $f_{\theta}$ used in the proposed ODE function is computed as follows:
\begin{equation}
    f_\theta(\boldsymbol{z}_{0})=(g(\boldsymbol{z}_{0}) + \mathbf{1}) \odot h(\boldsymbol{z}_{0})
 -\lambda\big(\boldsymbol{z}^{s}\big)\, \boldsymbol{z}_{0}, \quad \boldsymbol{z}_0 = \left[ \boldsymbol{z}^{s},\boldsymbol{z}^{g} \right],
\end{equation}
where $\odot$ represents the element-wise multiplication. Initially, the vector field is computed by the general residual block $h(\boldsymbol{z}_{0})$ and updated by a gated vector field with a sigmoid function $\sigma$ as:
\begin{equation}
g(\boldsymbol{z}_{0})=\sigma(W_g\boldsymbol{z}_{0}+\boldsymbol{b}_{g})\in(0,1)^{m+c},
\label{eq:gate}
\end{equation}
which provides state-adaptive modulation of the dynamics.
Finally, to regularize trajectories under noisy EEG inputs, we add an adaptive decay conditioned on the temporal stochastic state $\boldsymbol{z}^s$:
\begin{equation}
\lambda(\boldsymbol{z}^{s})=\texttt{Softplus}(W_a^{(2)} \circ\texttt{tanh}(W_s^{(1)}\boldsymbol{z}^{s} + \boldsymbol{b}^{1})+\boldsymbol{b}^{2})>0 .
\label{eq:decay-field}
\end{equation}

The latent trajectory $\boldsymbol{z}(t)$ at arbitrary time $t$ can be solved by:
\begin{equation}
\boldsymbol{z}_{t+K} = 
\begin{bmatrix} \boldsymbol{z}^{s}\\\boldsymbol{z}^{g}\end{bmatrix} + \int_{t+1}^{t+K}f_{\theta}\left( \begin{bmatrix} \boldsymbol{z}^{s}\\\boldsymbol{z}_{t}^{g}\end{bmatrix} ,t\right)dt \quad .
\label{ode2}
\end{equation}
The state solutions are calculated by solving with efficient numerical solvers in Figure \ref{fig:ST_ODE}(a), such as Runge-Kutta (RK) \citep{schober2019probabilistic}. 
The latent state at the next timestamp is updated as follows:
\begin{equation}
\boldsymbol{z}(t+\Delta t) = \boldsymbol{z}(t) + \frac{\Delta t}{6}(k_1 + 2k_2 + 2k_3 + k_4).
\end{equation}

\subsection{Graph Embedding Forecasting}
\label{subsec:graphforecasting}
Depending on the Eq. \ref{ode2}, the latent dynamic function and neural forecasting are presented as follow:
\begin{align}
    \{\boldsymbol{z}_{t+1}, \dots ,\boldsymbol{z}_{t+K} \} &= \texttt{ODESolver}\left(f_{\theta}, \left[ \boldsymbol{z}^{s}, \boldsymbol{z}^{g} \right] , [t+1,t+K]\right), \\
   \hat{\mathcal{G}}_{t+i} &= \mathrm{\Omega}(\boldsymbol{z}_{t+i}) \quad \forall i \in \{1,2, \dots, K\}, 
\end{align}
where the continuous latent trajectories $\{\boldsymbol{z}(t)\}_{t=1}^{K}$ are projected back to the future EEG node attributes with $\mathcal{V}$ the set of all possible unique nodes in $\mathcal{G}_{t+1:t+K}$ via a predictive module $\mathrm{\Omega: \mathbb{R}^{m+c} \mapsto \mathbb{R}^{d}}$, explicitly capturing spatial correlations across EEG channels over future $K$ time steps. Here, $\mathcal{X}_{:,>t} = [\mathcal{X}_{:,t+1}, \dots,\mathcal{X}_{:,t+K}]$ integrate all future node attributes.

Unlike the previous works, which focus on forecasting the temporal neural population dynamics.
Our learning objective is to predict the graph structure rather than the simple temporal dynamics, since neuron firing generally activates in the asynchronous channels simultaneously
    $\mathcal{L}_{\mathcal{G}} = \mathbb{E}_{\mathcal{G}} \left\| \hat{\mathcal{G}}_{t+1:K} - \mathcal{G}_{t+1:K} \right\|_{2}.$
We first train the model in an unsupervised manner using dynamic graph forecasting loss to capture continuous neural dynamics via ODE solvers. Then we pooling the latent continuous trajectory $z(t)$ extracted from the ODE solver with entire timesteps for downstream fine-tuning, like classification.

\section{Experiments}

In this section, we conduct experiments to answer the following research questions:  
\textbf{RQ1.} 
Does \method strengthen seizure detection capability through continuous forecasting on EEGs?
\textbf{RQ2.} 
How does the initial state $\vz_{0}$ affect the development  of latent neural trajectory? 
\textbf{RQ3.} 
Does our objective of $\mathrm{\Omega}$ facilitate dynamic optimization?
Details can be found in Appendix \ref{apdx:setting}.

\subsection{Experimental Setup}





\textbf{Tasks.} \quad In this study, we evaluate our \method for modeling neuronal population dynamics using seizure detection. 
Seizure detection is defined as a binary classification task that aims to distinguish between seizure and non-seizure EEG segments known as epochs. 
This task is fundamental to automated seizure monitoring systems.


\begin{wraptable}{r}{0.5\linewidth}  
  \vspace{-16pt}
  \centering
  \caption{Results (AUROC↑, F1↑) on \textbf{TUSZ} (12s and 60s seizure detection) against discrete and continuous baselines, with options on the gate and stochastic regularization. (-: w/o, +Random: gate with random coefficients for stochastic regularization.) Bold = best.}
  \label{tab:RQ2-2}
  \small
  \setlength{\tabcolsep}{2pt}
  \begin{tabular}{c l c c c}
    \toprule
    \multirow{2}{*}{Model} & Method & T(s) & AUROC & F1 \\
    \cmidrule(lr){2-5}
    \multirow{4}{*}{\rotatebox[origin=c]{90}{Discrete \& Continuous}}
      & BIOT    & 12  &  0.772$\pm$0.006 &  0.294$\pm$0.006\\
      &             & 60 & 0.642$\pm$0.009 & 0.256$\pm$0.003\\
      \cmidrule(lr){2-5}
      & DCRNN     & 12  & 0.816$\pm$0.002 & 0.416$\pm$0.009\\
      &             & 60 & 0.802$\pm$0.003 & 0.413$\pm$0.005\\
      \cmidrule(lr){2-5}
      & latent-ODE     & 12  & 0.791$\pm$0.004 & 0.385$\pm$0.005\\
      &             & 60 & 0.745$\pm$0.036 & 0.331$\pm$0.031\\
      \cmidrule(lr){2-5}
      & \method     & 12  & \bf 0.881$\pm$0.006 & \bf 0.496$\pm$0.017\\
      &             & 60 & \bf 0.828$\pm$0.003 & \bf 0.430$\pm$0.021\\
    \midrule
    \multirow{4}{*}{\rotatebox[origin=c]{90}{\method}}
      & - Gate    & 12  &  0.867$\pm$0.004 &  0.488$\pm$0.007\\
      &           & 60 &  0.821$\pm$0.034 &  0.424$\pm$0.003\\
      \cmidrule(lr){2-5}
      & - Stochastic     & 12  & 0.848$\pm$0.017 & 0.462$\pm$0.013\\
      &             & 60 & 0.817$\pm$0.029 & 0.414$\pm$0.047\\
      \cmidrule(lr){2-5}
     & +Random      & 12 & 0.860$\pm$0.017 & 0.474$\pm$0.033\\
      &             & 60 & 0.819$\pm$0.026 & 0.418$\pm$0.017\\
  \end{tabular}
\end{wraptable}

\textbf{Baseline methods.}
We select two baselines that study neural population dynamic studies: DCRNN \citep{li2017diffusion} which has a reconstruction objective. 
We also compare it against the benchmark Transformer BIOT \citep{yang2023biot}, which captures temporal-spatial information for EEG tasks. 
Finally, we compare against a standard baseline CNN-LSTM \citep{9175641}.

\begin{table*}[t]
\centering
\small
\setlength{\tabcolsep}{5pt}
\caption{Main results on \textbf{TUSZ} (12s seizure detection) and \textbf{TUAB}. 
\textbf{Bold} and \underline{underline} indicate best and second-best results.
$\star$: The performance depends on the discrete multi-steps forecasting. 
$\dag$: The performance depends on the \textit{continuous} multi-steps forecasting. 
$\ddag$: The performance depends on the \textit{continuous} single-step forecasting.
}
\label{tab:main_tusz_tuab}
\begin{threeparttable}
\resizebox{0.99\linewidth}{!}{
\begin{tabular}{l
S[table-format=1.3(2)] S[table-format=1.3(2)] S[table-format=1.3(2)] S[table-format=1.3(2)]
S[table-format=1.3(2)] S[table-format=1.3(2)]}
\toprule
\multirow{2}{*}{Method} 
& \multicolumn{3}{c}{\textbf{TUSZ}} & \multicolumn{3}{c}{\textbf{TUAB}} \\
\cmidrule(lr){2-4}\cmidrule(lr){5-7}
& {Acc} & {F1} & {AUROC}
& {Acc} & {F1} & {AUROC} \\
\midrule
CNN-LSTM      & 0.735(3) & 0.347(12)  & 0.757(3) & 0.741(2) & 0.736(7) & 0.813(3) \\
BIOT          & 0.702(3) & 0.294(6)  & 0.772(6) & 0.717(2) & 0.713(4) & 0.788(2) \\
EvolveGCN     &  0.769(2) & 0.385(5)  & 0.791(4) & 0.708(3) & 0.707(2) & 0.777(3) \\
DCRNN         & 0.816(2) & 0.416(9)  & 0.825(2) & 0.768(4) & 0.769(2)  & 0.848(2) \\
\midrule
latent-ODE  &  0.827(4) & 0.470(5)  & 0.849(4) & 0.749(3) & 0.745(2) & 0.829(4) \\
latent-ODE (RK4)  &  0.821(3) & 0.465(1)  & 0.845(4) & 0.746(2) & 0.739(2) & 0.823(3) \\
ODE-RNN  &  0.802(2) & 0.455(7)  & 0.855(3) & 0.751(3) & 0.744(4) & 0.838(5) \\
neural SDE  &  0.857(2) & 0.467(3)  & 0.851(2) & 0.768(3) & 0.751(3) & 0.834(2) \\
Graph ODE  &  0.849(3) & 0.475(5)  & 0.841(3) & 0.757(3) & 0.737(6) & 0.823(4) \\
$\method^\dag$  & \second{$0.869 \pm 0.003$} & \second{$0.488\pm 0.015$}  & $\second{0.875 \pm 0.005}$ & $\second{0.771 \pm 0.005 }$ & $\second{0.770 \pm 0.005 }$  & $\second{0.849 \pm 0.003 }$ \\

$\method^\ddag$ & $\best{0.877 \pm 0.004} $  & $\best{0.496 \pm 0.017} $   & $\best{0.881 \pm 0.006}$  &  $\best{0.778 \pm 0.003 }$ & $\best{0.774 \pm 0.005 }$  & $\best{0.857 \pm 0.005 }$\\

\bottomrule
\end{tabular}
}

\end{threeparttable}
\end{table*}

\textbf{Metrics.}
To answer \textbf{RQ1}, we evaluate the model using the Area Under the Receiver Operating Characteristic Curve (AUROC) and the F1 score. 
The AUROC measures the ability of the models across varying thresholds, while the F1 score highlights the balance between precision and recall at its optimal threshold for classification.
For \textbf{RQ2}, we measure the structural similarity of the predicted graph using the Global Jaccard Index (GJI) $
    \mathtt{GJI}(\mathcal{E}_{true},\mathcal{E}_{Pred}) = \frac{\lvert \mathcal{E}_{true}\cap \mathcal{E}_{Pred}\rvert}{\lvert \mathcal{E}_{true}\cup \mathcal{E}_{Pred}\rvert}$ \citep{castrillo2018dynamic}.
For \textbf{RQ3}, We compute the cosine similarity of predicted node embeddings.

\begin{figure*}[htbp]
\centering
\includegraphics[width = 0.99\linewidth]{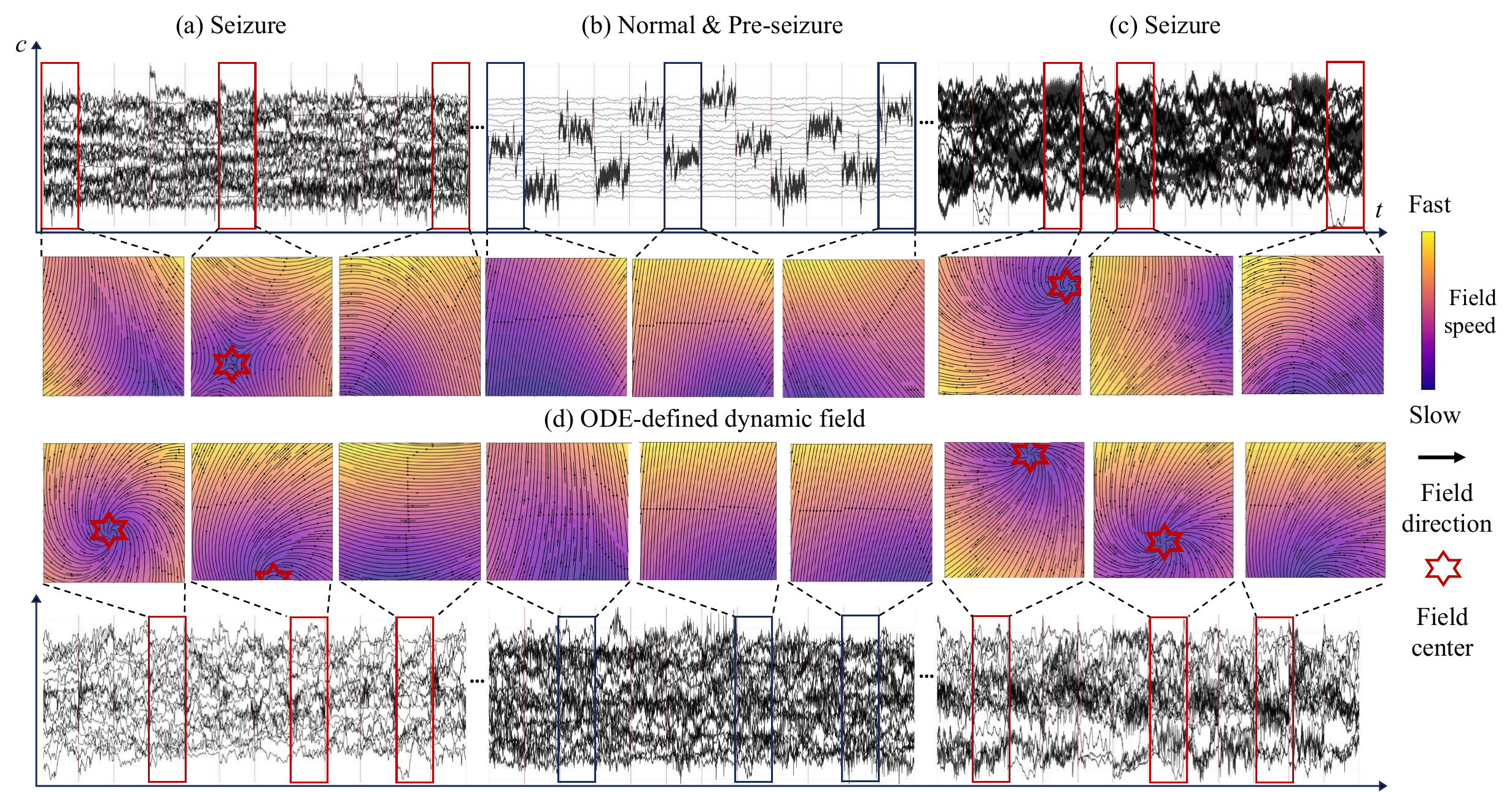}
\caption{Visualization results between the multichannel EEG signal (upper and lower) and its latent dynamic field $f_{\theta}$ (middle) obtained by \method.
Local minima appearing in (a) and (c) indicate rapid changes, corresponding to seizure states.
}
\label{fig:dynamic_field}
\end{figure*}


\subsection{Results}

\subsubsection{Main Result} 

\textbf{RQ1} concerns the continuous forecasting capability on EEG.
Table~\ref{tab:main_tusz_tuab} summarizes seizure detection accuracy across models on the TUSZ and TUAB datasets for a duration of 12 seconds.
Our \method consistently outperforms all baselines based on the AUROC and F1 score, demonstrating the superiorty of continuous forecasting. 
Notably, our single-step forecasting achieves an AUROC of $0.881\pm 0.006$ and an F1 score of $ 0.496 \pm 0.017 $, surpassing latent-ODE.
Our multi-step forecasting attains a recall of $0.563 \pm 0.015$, balancing overall detection capability and positive-instance coverage. 
These results indicate that \method is more effective in capturing the transient dynamics of EEGs than the fixed-time interval or reconstruction baselines.

\begin{figure*}[htbp]
\centering
\includegraphics[width = 0.99\linewidth]{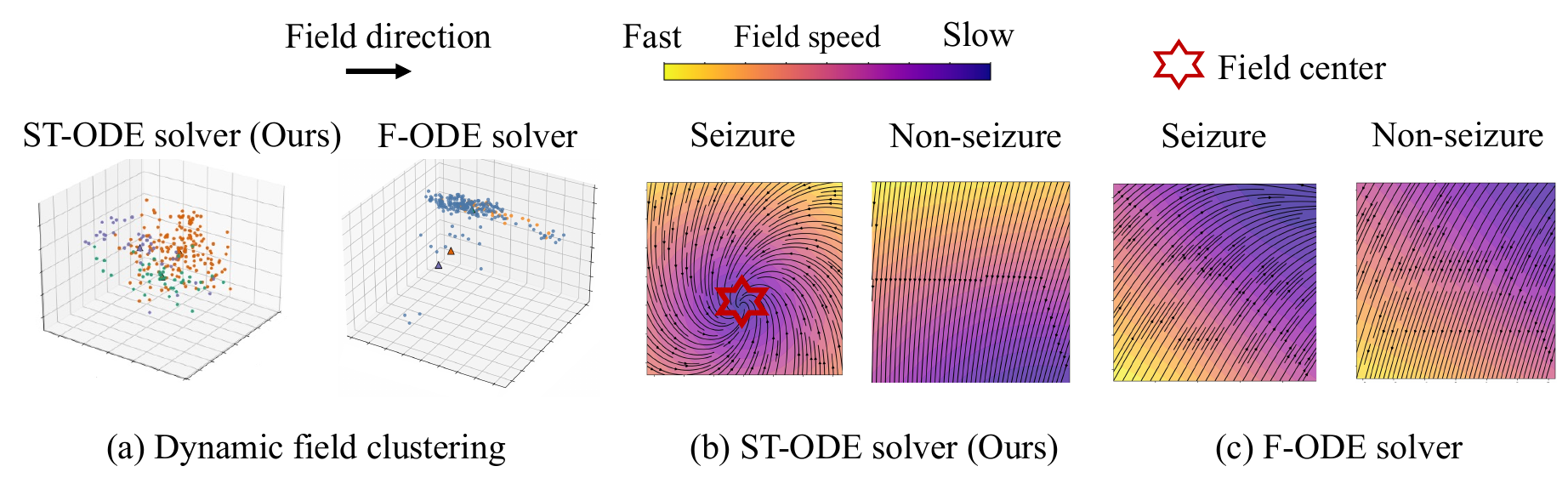}
\caption{Visualizing learned dynamic fields between our spatial-temporal(ST)-ODE solver and the frequency (F)-ODE solver.}
\vspace{-10pt}
\label{fig:ode_com}
\end{figure*}

To further illustrate this point, we visualize the dynamic field $f_{\theta}$ of the latent space in Fig.\@ \ref{fig:dynamic_field}. 
This dynamic field characterizes the difference between seizure and normal states.
This is most apparent from the centers in the seizure figures in Figure \ref{fig:dynamic_field}(a) and \ref{fig:dynamic_field}(c) while absent from the normal \& pre-seizure states \ref{fig:dynamic_field}(b).
These centers depict an area where gradients point and eventually the flows converge.
This aligns well with the corresponding EEGs that show wild-type oscillations with high frequency components.
In contrast, for the normal \& pre-seizure data, such centers are not present in the field, showing that the dynamics is driven mainly by low-frequency oscillations.
It is worth noting that such visualization is only available to continuous dynamics modeling of our method.

In summary, we can answer \textbf{RQ1} as follows: 
through continuous forecasting, \method outperforms existing baselines in terms of seizure detection by accurately depicting neural population dynamics. 
The learned field $f_{\theta}$ can clearly delineate the boundary between seizure and normal states via its vector field representation of neuronal activity.
Unlike discrete-time interval and reconstruction-based baselines, \method provides an arbitrary temporal resolution and it is sensitive to transient neural changes.
We have verified that it helps capture the transition process of different brain states.

\subsubsection{Dynamic graph forecasting evaluation}

\begin{wrapfigure}[32]{r}{0.45\textwidth}
\vspace{-4pt}
\centering
\includegraphics[width=0.99\linewidth]{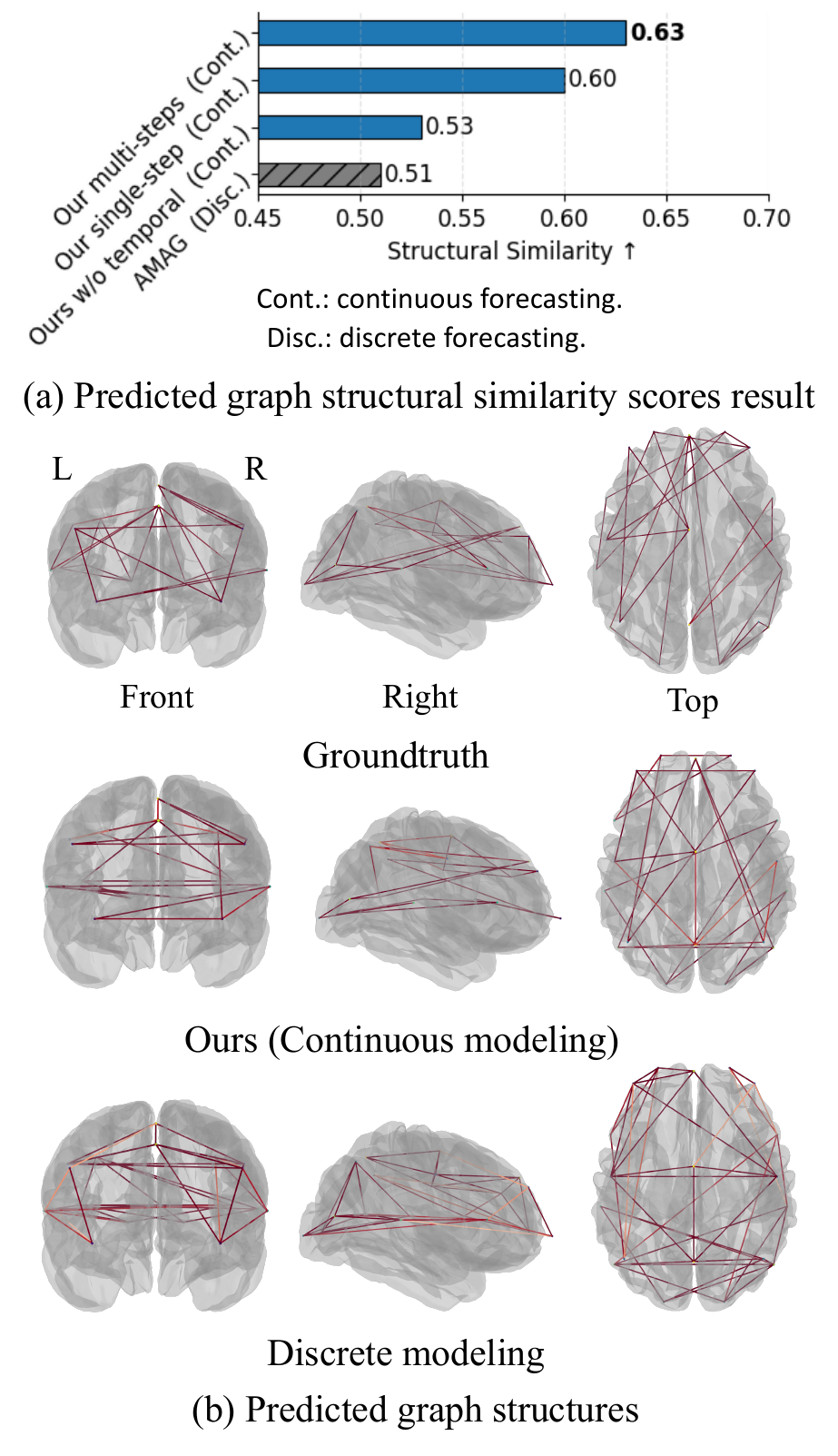}
\caption{Results on (a) graph similarity and (b) functional connections.}
\label{fig:brain_adj}
\end{wrapfigure}

\textbf{RQ2} concerns initial state $\vz_0$.
Fig.\@ \ref{fig:brain_adj} depicts the predicted connectivity patterns and edge densities.
It is evident that \method is closer to the ground truth than AMAG in terms of showing a more consistent topology. 
Consistent structural features with small offsets are crucial for correctly modeling brain dynamics. 
\method utilizes stocasticity in the raw EEG signal as an implicit regularization term.
This term helps enhance the generalization ability of continuous trajectory inference, as shown in Figure \ref{fig:brain_adj}(a), rising from $0.53$ to $0.63$ and maintains a consistent structure. 
We are ready to answer \textbf{RQ2}, given our $\vz_0$, \method can generate latent trajectories that respect EEG dynamics and maintain continuous evolutionary properties.
\captionsetup{skip=5pt}
Table~\ref{tab:RQ2-2} describes the seizure detection performance under $12$s and $60$s, comparing discrete and continuous baselines with \method. \method achieves the best or tied-best results at both horizons, indicating that adaptive vector field effectively strengthens stability. The ablations further validate our design by removing the gating mechanism leads to performance drop from $0.881$ to $0.867$, highlighting the adaptive vector field can achieve stable trajectory evolution. Removing stochastic regularization also degrades F1 from $0.496$ to $0.462$, proofing that stochastic regularization mitigates dynamics instability caused by noise. In contrast, using a gate with random coefficients for stochastic regularization still underperforms the full model, implying that our learnable regularization is more effective.

\textbf{RQ3} concerns consistency in the graphs.
Figure \@ \ref{fig:brain_adj} shows the effectiveness of our objective $\Omega$ that helps predict dynamic graph structures.
It is visible that \method achieves higher similarity scores ($0.53$ $\rightarrow$ $0.63$) than the discrete predictor, indicating that \method more accurately captures the true graph structure with the help of $\Omega$.
The similarity matrices reveal that ours aligns more closely in terms of local correlation distribution, in which the discrete predictor exhibits notable discrepancies in certain block structures. 
Now we can answer \textbf{RQ3}: the explicit graph embedding target improves forecasting accuracy.
This is achieved by guiding the vector field $f_{\theta}$ to learn continuous trajectories that align well with neural activity, leading to more reliable prediction.

\begin{figure*}[t]
\centering
\includegraphics[width = 0.99\linewidth]{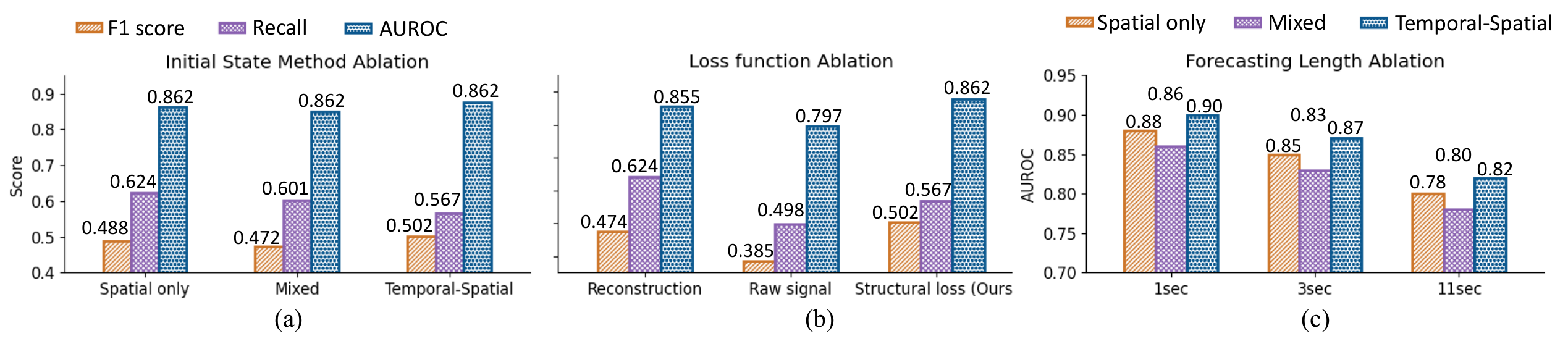}
\caption{ 
Summary of ablation study.
(a) State initialization. 
We compare spatial-only, mixed, and temporal–spatial initialization and summarized results in F1, Recall and AUROC.
Temporal–Spatial achieves the best F1 $(0.502)$ with a competitive recall.
(b) Loss function. Replacing our structural forecasting loss with reconstruction-only or raw-signal forecasting degrades performance on AUROC.
(c) Forecast horizon. AUROC decreases as the horizon grows ($1$s $\rightarrow 3$s $\rightarrow$ $11$s), and Temporal–Spatial remains the best across all horizons over others.}
\vspace{-3pt}
\label{fig:ablation}
\end{figure*}

\subsubsection{Ablation study} 

We perform ablation study on the following factors of \method:
initialization $\vz_0$, loss objective $\Omega$ and forecasting horizon, the results are summarized in Figure \ref{fig:ablation}.

\textbf{Initial state.} Temporal–spatial initial state option yields the best performance, achieving the highest AUROC $(0.877)$ and surpassing spatial-only $(0.862)$ and mix up $(0.851)$. It mitigates sensitivity to initial conditions and delivers the largest gains at the longest horizon ($11$s).
\textbf{Loss objective.} Our structural multi-step forecasting consistently outperforms reconstruction-only and raw-signal forecasting across F1/Recall/AUROC, indicating that geometry-aware regularization improves dynamical modeling.
We attribute the gains to \method that couples the spectral–spatial structure with EEG dynamics and enables more stable integration and stronger generalization.

\begin{table}[t]
  \captionsetup{skip=2pt}
  \centering
  \small

  \begin{minipage}[t]{0.48\linewidth}
    \captionof{table}{Computational cost with wall-clock time (s) and NFEs.}
    \label{tab:time}
    \setlength{\tabcolsep}{2pt}
    \begin{tabular}{c l c c c}
      \toprule
      \multirow{2}{*}{Type} & Model & Param. & Wall & NFEs \\
      \cmidrule(lr){2-5}
      \multirow{3}{*}{\rotatebox[origin=c]{90}{\scriptsize Discrete}}
        & CNN-LSTM  & 5976K  & 0.586$\pm$0.004 & -\\
        \cmidrule(lr){2-5}
        & BIOT      & 3174K  & 0.508$\pm$0.003 & -\\
        \cmidrule(lr){2-5}
        & DCRNN     & 281K   & 0.418$\pm$0.006 & -\\
        
      \midrule
      \multirow{4}{*}{\rotatebox[origin=c]{90}{\scriptsize Continuous}}
        & latent-ODE & 386K  & 0.421$\pm$0.002 & 102\\
        \cmidrule(lr){2-5}
        & ODE-RNN    & 675K  & 0.601$\pm$0.005 & 189\\
        \cmidrule(lr){2-5}
        & neural SDE & 346K  & 0.482$\pm$0.003 & 153\\
        \cmidrule(lr){2-5}
        & \method    & 459K  & 0.516$\pm$0.002 & 164\\
      \bottomrule
    \end{tabular}
  \end{minipage}\hfill
  \begin{minipage}[t]{0.50\linewidth}
    \captionof{table}{Ablation study on Top-$\tau{=}3$ and different regularizer options. Bold denotes the best.\\}
    \label{tab:ablation_top}
    \setlength{\tabcolsep}{2pt}
    \begin{tabular}{c l c c}
      \toprule
      \multirow{2}{*}{Model} & Regularizer & AUROC & Recall \\
      \cmidrule(lr){2-4}
      \multirow{3}{*}{\rotatebox[origin=c]{90}{\scriptsize latent-ODE}}
        & Shrinkage        & 0.833$\pm$0.032 & 0.567$\pm$0.021\\
      \cmidrule(lr){2-4}
        & Graphical lasso   & 0.846$\pm$0.025 & 0.557$\pm$0.022\\
      \cmidrule(lr){2-4}
        & Norm              & 0.849$\pm$0.004 & 0.575$\pm$0.005\\
      \midrule
      \multirow{3}{*}{\rotatebox[origin=c]{90}{\scriptsize \method}}
        & Shrinkage         & 0.872$\pm$0.023 & 0.606$\pm$0.035\\
      \cmidrule(lr){2-4}
        & Graphical lasso   & 0.872$\pm$0.017 & \textbf{0.613$\pm$0.033}\\
      \cmidrule(lr){2-4}
        & Norm              & \textbf{0.881$\pm$0.006} & 0.605$\pm$0.003\\
      \bottomrule
    \end{tabular}
  \end{minipage}
\end{table}

Table~\ref{tab:time} shows single-batch inference cost for discrete vs. continuous baselines, including parameters, wall-clock time, and NFEs (only for solver-based models). Discrete methods have fixed-depth computation, so latency mainly follows model size/sequence length. NFEs are shown only for the ODE solver-based models. \method contains $459$k parameters with $164$ NFEs, and $0.516$s per batch, which falls in the same latency band as discrete models with fixed-depth computation. These results indicate that \method does not introduce prohibitive cost in practice, and the reduced NFEs suggest a more stable integration than other complicated continuous baselines.

Table~\ref{tab:ablation_top} evaluates sensitivity to top-$\tau{=}3$ sparsity and regularization options. Adding regularization improves Recall, confirming that norm correlation graphs are noisy and susceptible to volume conduction, while regularized connectivity is more reliable. The performance is stable across regularization options. Concretely, an ODE solver can achieve better performance with sparser, regularized graphs. Graphical lasso or Norm with $3$ sparsity yields the best in both AUROC and Recall. For \method, Norm with $3$ sparsity achieves the best AUROC (0.881), and Graphical lasso gets the highest Recall ($0.613$),  demonstrating robust dependence on graph-construction choices.
\section{Conclusion}
In this work, we introduced \method, a novel continuous-time dynamic modeling framework for modeling EEGs, designed explicitly to overcome critical limitations associated with discrete-time recurrent approaches. 
By adopting a neural ODE-based approach with adaptive vector field strategy, our model effectively captures the continuous neural dynamics and spatial interactions in EEG data. Although \method models latent dynamics in continuous time, the inputs and supervision are still based on epoched segments, which limits long-term continuous modeling. And the generalization to other neurological disorders or cognitive tasks remains to be explored.

\clearpage

\section*{Acknowledgment}
This study was partly supported by JST PRESTO (JPMJPR24TB), CREST (JPMJCR24Q5), ASPIRE (JPMJAP2329), and Moonshot R\&D (JPMJMS2033-14). JSPS KAKENHI Grant-in-Aid for Scientific Research Number JP24K20778, JST CREST JPMJCR23M3, JST START JPMJST2553, JST CREST JPMJCR20C6, JST K Program JPMJKP25Y6, JST COI-NEXT JPMJPF2009, JST COI-NEXT JPMJPF2115, the Future Social Value Co-Creation Project - Osaka University.





\bibliography{iclr2026_conference}

@article{li2024amag,
  title={AMAG: Additive, Multiplicative and Adaptive Graph Neural Network For Forecasting Neuron Activity},
  author={Li, Jingyuan and Scholl, Leo and Le, Trung and Rajeswaran, Pavithra and Orsborn, Amy and Shlizerman, Eli},
  journal={Advances in Neural Information Processing Systems},
  volume={36},
  year={2024}
}

@article{li2017diffusion,
  title={Diffusion convolutional recurrent neural network: Data-driven traffic forecasting},
  author={Li, Yaguang and Yu, Rose and Shahabi, Cyrus and Liu, Yan},
  journal={arXiv preprint arXiv:1707.01926},
  year={2017}
}

@inproceedings{
delavari2024synapsnet,
title={SynapsNet: Enhancing Neuronal Population Dynamics Modeling via Learning Functional Connectivity},
author={Parsa Delavari and Ipek Oruc and Timothy H Murphy},
booktitle={The First Workshop on NeuroAI @ NeurIPS2024},
year={2024}
}

@inproceedings{
klotergens2025physiomeode,
title={Physiome-{ODE}: A Benchmark for Irregularly Sampled Multivariate Time-Series Forecasting Based on Biological {ODE}s},
author={Christian Kl{\"o}tergens and Vijaya Krishna Yalavarthi and Randolf Scholz and Maximilian Stubbemann and Stefan Born and Lars Schmidt-Thieme},
booktitle={The Thirteenth International Conference on Learning Representations},
year={2025}
}

@article{brain_dynamics,
author = {Rolls, Edmund T. and Cheng, Wei and Feng, Jianfeng},
title = {Brain dynamics: Synchronous peaks, functional connectivity, and its temporal variability},
journal = {Human Brain Mapping},
pages = {2790-2801},
year = {2021}
}

@inproceedings{
kotoge2025dynamic,
title={EvoBrain: Dynamic Multi-channel EEG Graph Modeling for Time-Evolving Brain Network},
author={Rikuto Kotoge and Zheng Chen and Tasuku Kimura and Yasuko Matsubara and Takufumi Yanagisawa and Haruhiko Kishima and Yasushi Sakurai},
year={2025},
booktitle={Thirty-Ninth Conference on Neural Information Processing Systems}
}

@inproceedings{GNN_AAAI23,
author = {Ho, Thi Kieu Khanh and Armanfard, Narges},
title = {Self-Supervised Learning for Anomalous Channel Detection in EEG Graphs: Application to Seizure Analysis},
year = {2023},
booktitle = {Proceedings of the AAAI Conference on
Artificial Intelligence},
pages = {7866-7874}
}

@inproceedings{GNN_ICLR22,
title={Self-Supervised Graph Neural Networks for Improved Electroencephalographic Seizure Analysis},
author={Siyi Tang and Jared Dunnmon and Khaled Kamal Saab and Xuan Zhang and Qianying Huang and Florian Dubost and Daniel Rubin and Christopher Lee-Messer},
booktitle={International Conference on Learning Representations},
year={2022}
}

@inproceedings{chenIJCAI,
  author  = {Chen, Zheng and Zhu, Lingwei and Yang, Ziwei and Zhang, Renyuan},
  year    = {2022},
  title   = {Multi-Tier Platform for Cognizing Massive Electroencephalogram},
  booktitle={IJCAI-22}, 
  pages   = {2464-2470}
}

@article{Jones2022computational,
author = {Jones, David and Lowe, V. and Graff-Radford, J. and Botha, Hugo and Barnard, L. and Wiepert, D. and Murphy, Matthew and Murray, Melissa and Senjem, Matthew and Gunter, Jeffrey and Wiste, H. and Boeve, B. and Knopman, D. and Petersen, Ronald and Jack, C.},
year = {2022},
pages = {1643},
title = {A computational model of neurodegeneration in Alzheimer’s disease},
journal = {Nature Communications}
}

@article{NatNeuroscience2021,
author = {Li, Adam and Huynh, Chester and Fitzgerald, Zachary and Cajigas, Iahn and Brusko, Damian and Jagid, Jonathan and Claudio, Angel and Kanner, Andres and Hopp, Jennifer and Chen, Stephanie and Haagensen, Jennifer and Johnson, Emily and Anderson, William and Crone, Nathan and Inati, Sara and Zaghloul, Kareem and Bulacio, Juan and Gonzalez-Martinez, Jorge and Sarma, Sridevi},
year = {2021},
pages = {1-10},
title = {Neural fragility as an EEG marker of the seizure onset zone},
volume = {24},
journal = {Nature Neuroscience}
}

@inproceedings{pradeepkumar2025TFM,
 title={Tokenizing Single-Channel EEG with Time-Frequency Motif Learning},
  author={Pradeepkumar, Jathurshan and Piao, Xihao and Chen, Zheng and Sun, Jimeng},
 booktitle={The Fourteenth International Conference on Learning Representations},
year={2026},
url={https://openreview.net/forum?id=2sPmWHZ8Ir}
}

@INPROCEEDINGS{kotoge2024splitsee,
author={Kotoge, Rikuto and Chen, Zheng and Kimura, Tasuku and Matsubara, Yasuko and Yanagisawa, Takufumi and Kishima, Haruhiko and Sakurai, Yasushi},
  booktitle={2024 IEEE International Conference on Data Mining}, 
  title={SplitSEE: A Splittable Self-supervised Framework for Single-Channel EEG Representation Learning}, 
  year={2024},
  pages={741-746}
}

@inproceedings{yang2022unsupervised,
  title={Unsupervised time-series representation learning with iterative bilinear temporal-spectral fusion},
  author={Yang, Ling and Hong, Shenda},
  booktitle={International conference on machine learning},
  pages={25038--25054},
  year={2022},
  organization={PMLR}
}

@inproceedings{chen2023two,
  title={A two-view EEG representation for brain cognition by composite temporal-spatial contrastive learning},
  author={Chen, Zheng and Zhu, Lingwei and Jia, Haohui and Matsubara, Takashi},
  booktitle={Proceedings of the 2023 SIAM International Conference on Data Mining (SDM)},
  pages={334--342},
  year={2023},
  organization={SIAM}
}

@article{rubanova2019latent,
  title={Latent ordinary differential equations for irregularly-sampled time series},
  author={Rubanova, Yulia and Chen, Ricky TQ and Duvenaud, David K},
  journal={Advances in neural information processing systems},
  volume={32},
  year={2019}
}

@article{chen2018neural,
  title={Neural ordinary differential equations},
  author={Chen, Ricky TQ and Rubanova, Yulia and Bettencourt, Jesse and Duvenaud, David K},
  journal={Advances in neural information processing systems},
  volume={31},
  year={2018}
}

@inproceedings{hwang2021climate,
  title={Climate modeling with neural diffusion equations},
  author={Hwang, Jeehyun and Choi, Jeongwhan and Choi, Hwangyong and Lee, Kookjin and Lee, Dongeun and Park, Noseong},
  booktitle={2021 IEEE International Conference on Data Mining (ICDM)},
  pages={230--239},
  year={2021},
  organization={IEEE}
}

@InProceedings{pmlr-v209-tang23a,
  title = 	 {Modeling Multivariate Biosignals With Graph Neural Networks and Structured State Space Models},
  author =       {Tang, Siyi and Dunnmon, Jared A and Liangqiong, Qu and Saab, Khaled K and Baykaner, Tina and Lee-Messer, Christopher and Rubin, Daniel L},
  booktitle = 	 {Proceedings of the Conference on Health, Inference, and Learning},
  pages = 	 {50--71},
  year = 	 {2023}
}

@inproceedings{fang2021spatial,
  title={Spatial-temporal graph ode networks for traffic flow forecasting},
  author={Fang, Zheng and Long, Qingqing and Song, Guojie and Xie, Kunqing},
  booktitle={Proceedings of the 27th ACM SIGKDD conference on knowledge discovery \& data mining},
  pages={364--373},
  year={2021}
}

@inproceedings{park2021vid,
  title={Vid-ode: Continuous-time video generation with neural ordinary differential equation},
  author={Park, Sunghyun and Kim, Kangyeol and Lee, Junsoo and Choo, Jaegul and Lee, Joonseok and Kim, Sookyung and Choi, Edward},
  booktitle={Proceedings of the AAAI Conference on Artificial Intelligence},
  volume={35},
  number={3},
  pages={2412--2422},
  year={2021}
}

@INPROCEEDINGS{10020662,
  author={Li, Alexis},
  booktitle={2022 IEEE International Conference on Big Data (Big Data)}, 
  title={BrainMixup: Data Augmentation for GNN-based Functional Brain Network Analysis}, 
  year={2022},
  volume={},
  number={},
  pages={4988-4992},
  doi={10.1109/BigData55660.2022.10020662}}

@inproceedings{kan2023r,
  title={R-mixup: Riemannian mixup for biological networks},
  author={Kan, Xuan and Li, Zimu and Cui, Hejie and Yu, Yue and Xu, Ran and Yu, Shaojun and Zhang, Zilong and Guo, Ying and Yang, Carl},
  booktitle={Proceedings of the 29th ACM SIGKDD Conference on Knowledge Discovery and Data Mining},
  pages={1073--1085},
  year={2023}
}

@INPROCEEDINGS{9630194,
  author={Demir, Andac and Koike-Akino, Toshiaki and Wang, Ye and Haruna, Masaki and Erdogmus, Deniz},
  booktitle={2021 43rd Annual International Conference of the IEEE Engineering in Medicine \& Biology Society (EMBC)}, 
  title={EEG-GNN: Graph Neural Networks for Classification of Electroencephalogram (EEG) Signals}, 
  year={2021},
  volume={},
  number={},
  pages={1061-1067},
  doi={10.1109/EMBC46164.2021.9630194}}

@inproceedings{gadgil2020spatio,
  title={Spatio-temporal graph convolution for resting-state fMRI analysis},
  author={Gadgil, Soham and Zhao, Qingyu and Pfefferbaum, Adolf and Sullivan, Edith V and Adeli, Ehsan and Pohl, Kilian M},
  booktitle={Medical Image Computing and Computer Assisted Intervention--MICCAI 2020: 23rd International Conference, Lima, Peru, October 4--8, 2020, Proceedings, Part VII 23},
  pages={528--538},
  year={2020},
  organization={Springer}
}

@inproceedings{yang2023biot,
    title={BIOT: Biosignal Transformer for Cross-data Learning in the Wild},
    author={Yang, Chaoqi and Westover, M Brandon and Sun, Jimeng},
    booktitle={Thirty-seventh Conference on Neural Information Processing Systems},
    year={2023},
    url={https://openreview.net/forum?id=c2LZyTyddi}
}

@article{shah2018temple,
  title={The temple university hospital seizure detection corpus},
  author={Shah, Vinit and Von Weltin, Eva and Lopez, Silvia and McHugh, James Riley and Veloso, Lillian and Golmohammadi, Meysam and Obeid, Iyad and Picone, Joseph},
  journal={Frontiers in neuroinformatics},
  volume={12},
  pages={83},
  year={2018},
  publisher={Frontiers Media SA}
}

@article{SODorAAAI2025,
      title={Long-Term EEG Partitioning for Seizure Onset Detection}, 
journal={Proceedings of the AAAI Conference on Artificial Intelligence}, 
author={Chen, Zheng and Matsubara, Yasuko and Sakurai, Yasushi and Sun, Jimeng}, 
year={2025}, 
pages={14221-14229}  
}

@INPROCEEDINGS{9175641,
  author={Ahmedt-Aristizabal, David and Fernando, Tharindu and Denman, Simon and Petersson, Lars and Aburn, Matthew J. and Fookes, Clinton},
  booktitle={2020 42nd Annual International Conference of the IEEE Engineering in Medicine \& Biology Society (EMBC)}, 
  title={Neural Memory Networks for Seizure Type Classification}, 
  year={2020},
  volume={},
  number={},
  pages={569-575},
  doi={10.1109/EMBC44109.2020.9175641}
}

@article{schober2019probabilistic,
  title={A probabilistic model for the numerical solution of initial value problems},
  author={Schober, Michael and S{\"a}rkk{\"a}, Simo and Hennig, Philipp},
  journal={Statistics and Computing},
  volume={29},
  number={1},
  pages={99--122},
  year={2019},
  publisher={Springer}
}

@inproceedings{kim2021inferring,
  title={Inferring latent dynamics underlying neural population activity via neural differential equations},
  author={Kim, Timothy D and Luo, Thomas Z and Pillow, Jonathan W and Brody, Carlos D},
  booktitle={International Conference on Machine Learning},
  pages={5551--5561},
  year={2021},
  organization={PMLR}
}

@INPROCEEDINGS{10230734,
  author={Cai, Hongmin and Dan, Tingting and Huang, Zhuobin and Wu, Guorong},
  booktitle={2023 IEEE 20th International Symposium on Biomedical Imaging (ISBI)}, 
  title={OSR-NET: Ordinary Differential Equation-Based Brain State Recognition Neural Network}, 
  year={2023},
  volume={},
  number={},
  pages={1-5},
  doi={10.1109/ISBI53787.2023.10230734}}

@inproceedings{han2024brainode,
  title={Brainode: Dynamic brain signal analysis via graph-aided neural ordinary differential equations},
  author={Han, Kaiqiao and Yang, Yi and Huang, Zijie and Kan, Xuan and Guo, Ying and Yang, Yang and He, Lifang and Zhan, Liang and Sun, Yizhou and Wang, Wei and others},
  booktitle={2024 IEEE EMBS International Conference on Biomedical and Health Informatics (BHI)},
  pages={1--8},
  year={2024},
  organization={IEEE}
}

@article{hu2024modeling,
  title={Modeling latent neural dynamics with gaussian process switching linear dynamical systems},
  author={Hu, Amber and Zoltowski, David and Nair, Aditya and Anderson, David and Duncker, Lea and Linderman, Scott},
  journal={Advances in Neural Information Processing Systems},
  volume={37},
  pages={33805--33835},
  year={2024}
}

@article{vyas2020computation,
  title={Computation through neural population dynamics},
  author={Vyas, Saurabh and Golub, Matthew D and Sussillo, David and Shenoy, Krishna V},
  journal={Annual review of neuroscience},
  volume={43},
  number={1},
  pages={249--275},
  year={2020},
  publisher={Annual Reviews}
}

@article{churchland2012neural,
  title={Neural population dynamics during reaching},
  author={Churchland, Mark M and Cunningham, John P and Kaufman, Matthew T and Foster, Justin D and Nuyujukian, Paul and Ryu, Stephen I and Shenoy, Krishna V},
  journal={Nature},
  volume={487},
  number={7405},
  pages={51--56},
  year={2012},
  publisher={Nature Publishing Group UK London}
}

@article{mante2013context,
  title={Context-dependent computation by recurrent dynamics in prefrontal cortex},
  author={Mante, Valerio and Sussillo, David and Shenoy, Krishna V and Newsome, William T},
  journal={nature},
  volume={503},
  number={7474},
  pages={78--84},
  year={2013},
  publisher={Nature Publishing Group UK London}
}

@article{kingma2014adam,
  title={Adam: A method for stochastic optimization},
  author={Kingma, Diederik P},
  journal={arXiv preprint arXiv:1412.6980},
  year={2014}
}

@article{castrillo2018dynamic,
  title={Dynamic structural similarity on graphs},
  author={Castrillo, Eduar and Le{\'o}n, Elizabeth and G{\'o}mez, Jonatan},
  journal={arXiv preprint arXiv:1805.01419},
  year={2018}
}

@inproceedings{lu2018beyond,
  title={Beyond finite layer neural networks: Bridging deep architectures and numerical differential equations},
  author={Lu, Yiping and Zhong, Aoxiao and Li, Quanzheng and Dong, Bin},
  booktitle={International conference on machine learning},
  pages={3276--3285},
  year={2018},
  organization={PMLR}
}

@article{goldental2014computational,
  title={A computational paradigm for dynamic logic-gates in neuronal activity},
  author={Goldental, Amir and Guberman, Shoshana and Vardi, Roni and Kanter, Ido},
  journal={Frontiers in computational neuroscience},
  volume={8},
  pages={52},
  year={2014},
  publisher={Frontiers Media SA}
}

@article{oh2024stable,
  title={Stable neural stochastic differential equations in analyzing irregular time series data},
  author={Oh, YongKyung and Lim, Dong-Young and Kim, Sungil},
  journal={arXiv preprint arXiv:2402.14989},
  year={2024}
}

@article{pijn1997nonlinear,
  title={Nonlinear dynamics of epileptic seizures on basis of intracranial EEG recordings},
  author={Pijn, Jan Pieter M and Velis, Demetrios N and van der Heyden, Marcel J and DeGoede, Jaap and van Veelen, Cees WM and Lopes da Silva, Fernando H},
  journal={Brain topography},
  volume={9},
  number={4},
  pages={249--270},
  year={1997},
  publisher={Springer}
}

@inproceedings{xue2016minimum,
  title={Minimum number of sensors to ensure observability of physiological systems: A case study},
  author={Xue, Yuankun and Pequito, Sergio and Coelho, Joana R and Bogdan, Paul and Pappas, George J},
  booktitle={2016 54th Annual Allerton Conference on Communication, Control, and Computing (Allerton)},
  pages={1181--1188},
  year={2016},
  organization={IEEE}
}

@article{lehnertz2003seizure,
  title={Seizure prediction by nonlinear EEG analysis},
  author={Lehnertz, Klaus and Mormann, Florian and Kreuz, Thomas and Andrzejak, Ralph G and Rieke, Christoph and David, Peter and Elger, Christian E},
  journal={IEEE Engineering in Medicine and Biology Magazine},
  volume={22},
  number={1},
  pages={57--63},
  year={2003},
  publisher={IEEE}
}

@article{lehnertz2008epilepsy,
  title={Epilepsy and nonlinear dynamics},
  author={Lehnertz, Klaus},
  journal={Journal of biological physics},
  volume={34},
  number={3},
  pages={253--266},
  year={2008},
  publisher={Springer}
}

@article{mercier2024value,
  title={The value of linear and non-linear quantitative EEG analysis in paediatric epilepsy surgery: a machine learning approach},
  author={Mercier, Mattia and Pepi, Chiara and Carfi-Pavia, Giusy and De Benedictis, Alessandro and Espagnet, Maria Camilla Rossi and Pirani, Greta and Vigevano, Federico and Marras, Carlo Efisio and Specchio, Nicola and De Palma, Luca},
  journal={Scientific reports},
  volume={14},
  number={1},
  pages={10887},
  year={2024},
  publisher={Nature Publishing Group UK London}
}

@inproceedings{gupta2018re,
  title={Re-thinking EEG-based non-invasive brain interfaces: Modeling and analysis},
  author={Gupta, Gaurav and Pequito, S{\'e}rgio and Bogdan, Paul},
  booktitle={2018 ACM/IEEE 9th International Conference on Cyber-Physical Systems (ICCPS)},
  pages={275--286},
  year={2018},
  organization={IEEE}
}

@article{tzoumas2018selecting,
  title={Selecting sensors in biological fractional-order systems},
  author={Tzoumas, Vasileios and Xue, Yuankun and Pequito, S{\'e}rgio and Bogdan, Paul and Pappas, George J},
  journal={IEEE Transactions on Control of Network Systems},
  volume={5},
  number={2},
  pages={709--721},
  year={2018},
  publisher={IEEE}
}

@article{lu2021detection,
  title={Detection and classification of epileptic EEG signals by the methods of nonlinear dynamics},
  author={Lu, XiaoJie and Zhang, JiQian and Huang, ShouFang and Lu, Jun and Ye, MingQuan and Wang, MaoSheng},
  journal={Chaos, Solitons \& Fractals},
  volume={151},
  pages={111032},
  year={2021},
  publisher={Elsevier}
}

@article{martis2015epileptic,
  title={Epileptic EEG classification using nonlinear parameters on different frequency bands},
  author={Martis, Roshan Joy and Tan, Jen Hong and Chua, Chua Kuang and Loon, Too Cheah and YEO, SHARON WAN JIE and Tong, Louis},
  journal={Journal of Mechanics in Medicine and Biology},
  volume={15},
  number={03},
  pages={1550040},
  year={2015},
  publisher={World Scientific}
}

@article{lepeu2024critical,
  title={The critical dynamics of hippocampal seizures},
  author={Lepeu, Gregory and van Maren, Ellen and Slabeva, Kristina and Friedrichs-Maeder, Cecilia and Fuchs, Markus and Z’Graggen, Werner J and Pollo, Claudio and Schindler, Kaspar A and Adamantidis, Antoine and Proix, Timoth{\'e}e and others},
  journal={Nature communications},
  volume={15},
  number={1},
  pages={6945},
  year={2024},
  publisher={Nature Publishing Group UK London}
}

@inproceedings{gupta2019learning,
  title={Learning latent fractional dynamics with unknown unknowns},
  author={Gupta, Gaurav and Pequito, S{\'e}rgio and Bogdan, Paul},
  booktitle={2019 American Control Conference (ACC)},
  pages={217--222},
  year={2019},
  organization={IEEE}
}

@inproceedings{gupta2018dealing,
  title={Dealing with unknown unknowns: Identification and selection of minimal sensing for fractional dynamics with unknown inputs},
  author={Gupta, Gaurav and Pequito, S{\'e}rgio and Bogdan, Paul},
  booktitle={2018 Annual American Control Conference (ACC)},
  pages={2814--2820},
  year={2018},
  organization={IEEE}
}

@inproceedings{yang2019data,
  title={Data-driven perception of neuron point process with unknown unknowns},
  author={Yang, Ruochen and Gupta, Gaurav and Bogdan, Paul},
  booktitle={Proceedings of the 10th ACM/IEEE International Conference on Cyber-Physical Systems},
  pages={259--269},
  year={2019}
}

@article{yang2025spiking,
  title={Spiking dynamics of individual neurons reflect changes in the structure and function of neuronal networks},
  author={Yang, Ruochen and Ping, Heng and Xiao, Xiongye and Kiani, Roozbeh and Bogdan, Paul},
  journal={Nature Communications},
  volume={16},
  number={1},
  pages={6994},
  year={2025},
  publisher={Nature Publishing Group UK London}
}

@article{chen2024eeg,
  title={EEG emotion recognition based on ordinary differential equation graph convolutional networks and dynamic time wrapping},
  author={Chen, Yiyuan and Xu, Xiaodong and Bian, Xiaoyi and Qin, Xiaowei},
  journal={Applied Soft Computing},
  volume={152},
  pages={111181},
  year={2024},
  publisher={Elsevier}
}
\bibliographystyle{iclr2026_conference}

\clearpage
\appendix
\section{Experimental Settings}\label{apdx:setting}

\subsection{Discussion: Key insights of \method}
Conceptually, the major gain of our work comes from explicitly modeling continuous dynamics over graph structures. By capturing the dynamic evolution of EEG signals, the model can effectively handle substantial noise, randomness, and fluctuations. Our comparison with the baseline without continuous dynamics (i.e., using only a temporal GNN backbone) clearly supports this observation. Methodologically, our improvements arise from two key aspects: (i) obtaining a high-quality initialization \( \mathbf{z}_0 \), and (ii) formulating a vector field \( f_\theta \) that captures informative and stable dynamics. First, the reverse initial encoding provides a high-quality continuous representation that enables the model to unfold temporal information embedded in EEGs. This is achieved through a dual-encoder architecture that integrates spectral graph features with stochastic temporal signals. Second, the temporal–spatial ODE solver \( f_\theta \) incorporates the initialization into additive and gating operations, enabling adaptive emphasis on informative EEG connectivity patterns that encode richer dynamics (new Figure~xx in the revised manuscript). Furthermore, the stochastic regularizer mitigates the classical error-accumulation problem of ODEs by modeling stochasticity in the EEG time domain, thereby improving long-term stability. We also include a new ablation table (Table~2) to validate the contribution of each component and support the above points.

\subsection{Dynamic Spectral Graph Structure}\label{sec:graph}

Raw EEG signals consist of complicated neural activities overlapping in multiple frequency bands, each potentially encoding different functional neural dynamics. 
Directly analyzing EEG signals in the time domain often misses subtle state transitions occurring uniquely within specific frequency bands \citep{yang2022unsupervised,chen2023two}. 
Hence, it is beneficial to represent the intensity variations of frequency bands and waveforms by decomposing raw EEG signals into frequency components.
To effectively provide detailed insights for subtle state transitions, we perform the short-time Fourier transform (STFT) to each EEG epoch, preserving their non-negative log-spectral. Consequently, the multi-channel EEG recordings are processed as:
\begin{equation}
     \mathbf{X}_{t} = \sum_{t= \infty}^{-\infty}x[t] \,\omega[t-m]e^{-jwt},
\end{equation}
and a sequence of EEG epochs with their spectral representation is formulated as $\mathbf{X} \in \mathbb{R}^{N \times d \times T}$.

We then apply a graph representation by measuring the similarity among spectral representation $\mathbf{X}$ across EEG channels. Specifically, we define an adjacency matrix $\mathcal{A}_{t}(i,j)$ at each epoch $t$ as follows:
$\mathcal{A}_{t}(i,j) = \text{sim}(\mathbf{X}_{i,t},\mathbf{X}_{j,t})$
 and compute the normalized correlation between nodes $v_{i}$ and $v_{j}$, where the graph structure and its associated edge weight matrix $A_{i,j}$ are inferred from $X_{t}$ on for each $t$-th epoch. 
 We only preserve the top-$\tau$ highest correlations to construct the evident graphs without redundancy. To avoid redundant connections and clearly represent dominant spatial structures, we retain only the top-$\tau$ strongest connections at each epoch for sparse and meaningful graph representations.
Thus, we obtain a temporal sequence of EEG spectral graphs $\{{G}_t=(\mathcal{V}_t,\mathcal{A}_t)\}_{t=0}^{T}$.

\textbf{Temporal Graph Representation.}
Consider an EEG $\mathbf{X}$ consisting of $N$ channels and $T$ time points, we represent $\mathbf{X}$ as a graph, denoted as $\mathcal{G} = \{ \mathcal{V}, \mathcal{A}, \mathbf{X} \}$, where $\mathcal{V} = \{ v_1, \dots, v_N \}$ represents the set of nodes. 
Each node corresponds to an EEG channel. 
The adjacency matrix $\mathcal{A} \in \mathbb{R}^{N \times N \times T}$ encodes the connectivity between these nodes over time, with each element $a_{i,j,t}$ indicating the strength of connectivity between nodes $v_i$ and $v_j$ at the time point $t$.
Here, we redefine $T$ as a sequence of EEG segments, termed ``epochs'', obtained using a moving window approach. 
The embedding of node $v_i$ at the $t$-th epoch is represented as $h_{i,t} \in \mathbb{R}^m$.
Specifically, we perform the short-time Fourier transform (STFT) to each EEG epoch, referring to \citep{GNN_ICLR22}.
Then we measure the similarity among the spectral representation of the EEG channels to initial the $\mathcal{A}_{t}(i,j)$ for each epoch $t$.

\subsection{Datasets and Evaluation Protocols}
\textbf{Tasks.} \quad In this study, we evaluate our \method for modeling the neuronal population dynamics with the \textbf{seizure detection}. 
Seizure detection is defined as a binary classification task that aims to distinguish between seizure and non-seizure EEG segments known as epochs. 
This task is fundamental to automated seizure monitoring systems.

\textbf{Baseline methods.}
We select two baselines that study neural population dynamic studies: DCRNN \citep{li2017diffusion} that has a reconstruction objective; AMAG \citep{li2024amag} that has a discrete forecasting objective. 
We also compare against the benchmark Transformer BIOT \citep{yang2023biot} that captures temporal-spatial information for EEG tasks. 
Finally, we compare against a standard baseline CNN-LSTM \citep{9175641}.

\textbf{Datasets.}
We use the Temple University Hospital EEG Seizure (TUSZ) and the TUH Abnormal EEG Corpus (TUAB) \citep{shah2018temple}, the largest publicly available EEG seizure database. TUSZ contains 5,612 EEG recordings with 3,050 annotated seizures. Each recording consists of 19 EEG channels following the 10-20 system, ensuring clinical relevance. 
A key strength of TUSZ lies in its diversity, as the dataset includes data collected over different time periods, using various equipment, and covering a wide age range of subjects.
To provide normal controls, we sample studies from the “normal” subset of TUAB. Unless stated otherwise, recordings are processed with the same pipeline across corpora (canonical 10–20 montage with 19 channels and unified resampling), ensuring consistent preprocessing for cross-dataset evaluation.

\textbf{Metrics.}
To answer \textbf{RQ1}, we evaluate the model using the Area Under the Receiver Operating Characteristic Curve (AUROC) and the F1 score. 
AUROC measures the ability of models across varying thresholds, while the F1 score highlights the balance between precision and recall at its optimal threshold for classification.
For \textbf{RQ2}, we measure the predicted graph structural similarity using the Global Jaccard Index (GJI) \citep{castrillo2018dynamic}:
\begin{equation}
    \mathtt{GJI}(\mathcal{E}_{true},\mathcal{E}_{Pred}) = \frac{\lvert \mathcal{E}_{true}\cap \mathcal{E}_{Pred}\rvert}{\lvert \mathcal{E}_{true}\cup \mathcal{E}_{Pred}\rvert} \quad \quad .
\end{equation}

\textbf{Model training.} 
All models are optimized using the Adam optimizer \citep{kingma2014adam} with an initial learning rate of $1 \times 10^{-3}$ in the PyTorch and PyTorch Geometric libraries on NVIDIA A6000 GPU and AMD EPYC 7302 CPU.
We adopt the adaptive Runge-Kutta NODE integration solver (RK45) with relative tolerance set to $1 \times 10^{-5}$ for training.

\subsection{Hyperparameters}
All experiments are conducted on the TUSZ and TUAB dataset using CUDA devices and a fixed random seed of 123. EEG signals are preprocessed via Fourier transform, segmented into 12-second sequences with a 1-second step size, and represented as dynamic graphs comprising 19 nodes (EEG channels). Graph sparsification is achieved with Top-$k=3$ neighbors. Both dynamic and individual graphs use dual random-walk filters, whereas the combined graph employs a Laplacian filter.
\begin{figure}[!ht]
\centering
\includegraphics[width = 0.9\linewidth]{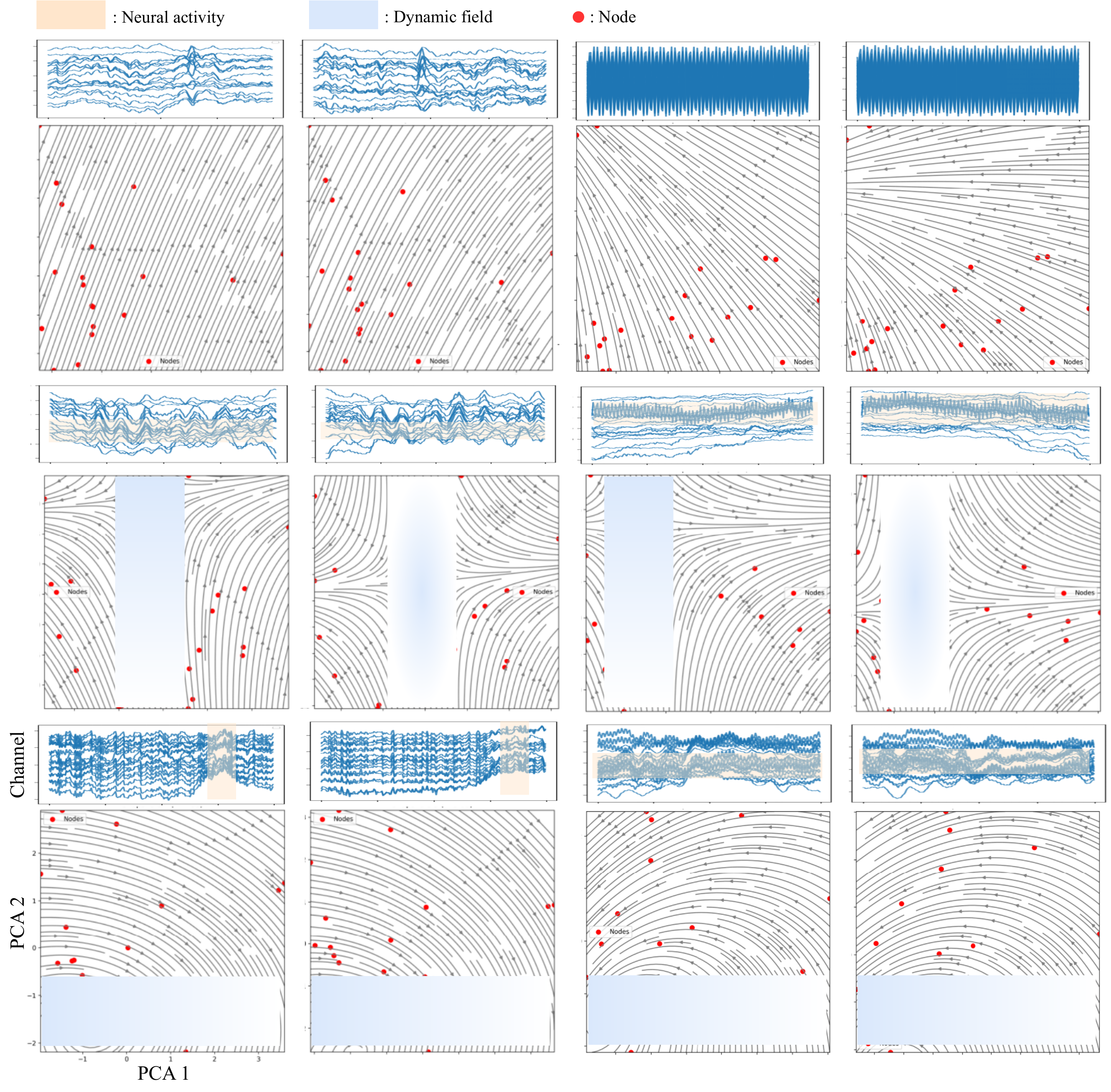}
\caption{Visualization results between the multichannel EEG signal (upper) and its latent dynamic field $f_{\theta}$ (lower) in our temporal-spatial neural ODE.}
\label{fig:DF}
\end{figure}
The default backbone is GRU-GCN for reverse initial state encoding, consisting of 2-layer GRU with 64 hidden units per layer. We also apply a CNN encoder with 3 hidden layers to extract the stochastic feature $z^{s}$ to obtain the final initial value $z_{0}$.
The convolution adopts a $2\times2$ kernel size with batch normalization and max pooling . Input and output feature dimensions are both 100, with the number of classes set to 1 for detection/classification tasks.

We train models using an initial learning rate of 3e-4, weight decay 5e-4, dropout rate 0.0, batch sizes of 128 (training) and 256 (validation/test), and a maximum of 100 epochs. Gradient clipping with a maximum norm of 5.0 and early stopping with a patience of 5 epochs are applied. Model checkpoints are selected by maximizing AUROC on the validation set (weighted averaging). When the metric is loss, we instead minimize it; all other metrics (e.g., F1, ACC) are maximized. Data augmentation is enabled by default, while curriculum learning is disabled unless otherwise stated.

\section{Additional Results}

Fig.\@ \ref{fig:DF} shows the visualization of the dynamic field $f_{\theta}$ of the latent space. It reveals distinct neural activity patterns: during synchronous low-frequency oscillations, dynamic field appears steady state, while high-frequency bursts trigger localized positive gradients, driving system activation. Asynchronous cross-channel interactions manifest as vortex-like flows, reflecting dynamic balance. Notably, continuous dynamic evolution offers finer temporal resolution at arbitrate time. \method enables early detection of neural transitions, better than discrete-time methods.

\begin{figure*}[!htbp]
    \subfigure[Comparison among groundtruth, graph output of our continuous predictor, graph output of discrete predictor.]{
        \includegraphics[width=0.9\linewidth]{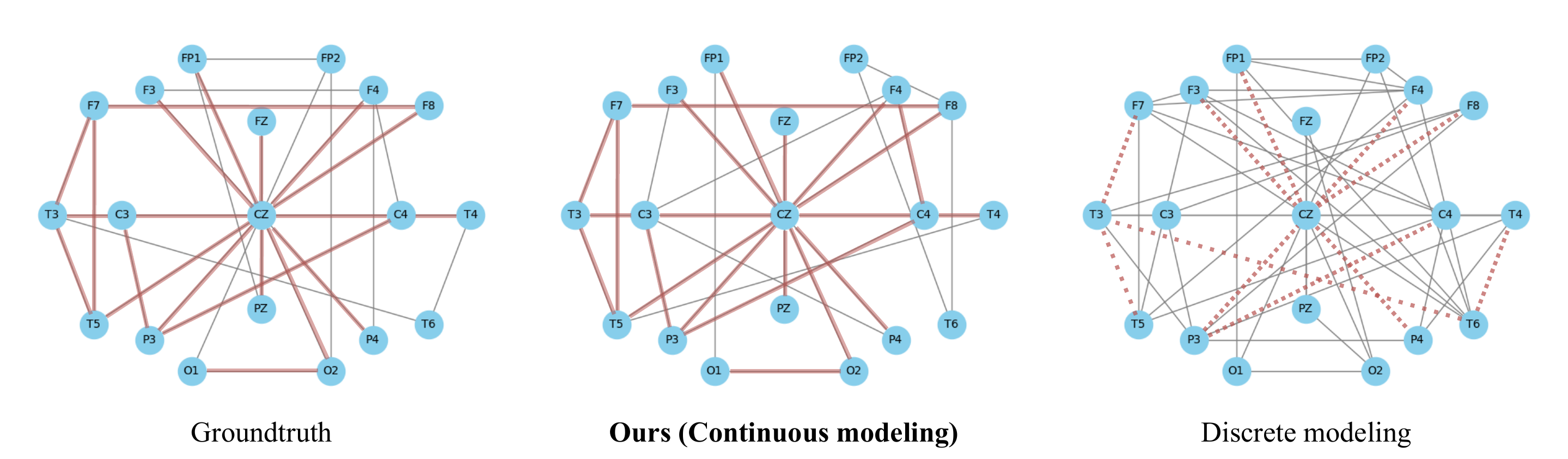}
        \label{fig:subfig1-b}
    }

    \vspace{1em}
    \centering
    \subfigure[Comparison of correlation scores between graph output of our continuous predictor, and graph output of discrete predictor.]{
        \includegraphics[width=0.8\linewidth]{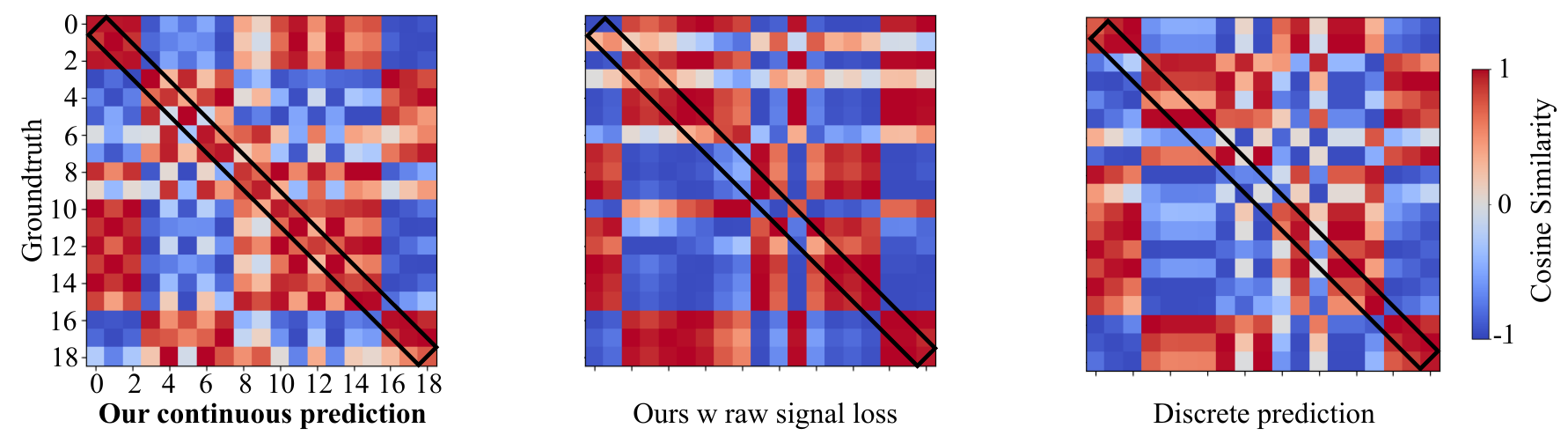}
        \label{fig:subfig1-a}
    }

    \caption{Comparison of the predicted graph output between our continuous predictor and discrete predictor.}
    \label{fig:subfig 1}
\end{figure*}

 Fig.\@ \ref{fig:subfig1-b} depicts the predicted connectivity patterns and edge densities from \method closer to the true connectivity than discrete predictor-based AMAG, leading to a significant topology consistency. These structural features are crucial for modeling consistent brain dynamics, as small topological offsets lead to correct brain activity for downstream tasks. The stochastic components of the raw EEG signal can be regard as an implicit regularity term, which helps to enhance the generalization ability of continuous trajectory inference and maintains consistency with the structure. The latent variable trajectories generated by \method not only maintain the continuous evolutionary properties, but also enhance the predictive ability of spatial consistency.

Fig.\@ \ref{fig:subfig 1} shows the effectiveness of predicting the dynamic graph structure depending on our meaningful forecasting objective $\Omega$. Fig.\@ \ref{fig:subfig1-a} present that \method can achieve higher similarity than the discrete predictor, indicating that the continuous prediction model more accurately captures the true graph structure. The similarity matrices reveal that ours aligns more closely in terms of local correlation distribution, in which the discrete predictor exhibits notable discrepancies in certain block structures. 
The explicit graph embedding target improves the forecasting accuracy, while effectively guides the vector field $f{\theta}$ to learn continuous trajectories aligned with the neural activity, leading to more reliable prediction.

\begin{table*}[t]
\centering
\small
\setlength{\tabcolsep}{5pt}
\caption{Ablation of pooling options over ODE-trajectory on \textbf{TUSZ} (12s seizure detection) and \textbf{TUAB}. 
\textbf{Bold} indicates best result.}
\label{tab:ablation_pool}
\begin{threeparttable}
\resizebox{0.99\linewidth}{!}{
\begin{tabular}{l
S[table-format=1.3(2)] S[table-format=1.3(2)] S[table-format=1.3(2)] S[table-format=1.3(2)]
S[table-format=1.3(2)] S[table-format=1.3(2)]}
\toprule
\multirow{2}{*}{Method} 
& \multicolumn{3}{c}{\textbf{TUSZ}} & \multicolumn{3}{c}{\textbf{TUAB}} \\
\cmidrule(lr){2-4}\cmidrule(lr){5-7}
& {Acc} & {F1} & {AUROC}
& {Acc} & {F1} & {AUROC} \\
\midrule
Max pooling& $\best{0.877 \pm 0.004} $  & $\best{0.496 \pm 0.017} $   & $\best{0.881 \pm 0.006}$  &  $\best{0.778 \pm 0.003 }$ & $\best{0.774 \pm 0.005 }$  & $\best{0.857 \pm 0.005 }$\\

Mean pooling  &  0.842(2) & 0.385(5)  & 0.827(3) & 0.748(2) & 0.635(2) & 0.827(4) \\

Sum pooling &  0.851(2) & 0.466(5)  & 0.867(4) & 0.753(3) & 0.755(2) & 0.831(4) \\

\bottomrule
\end{tabular}
}

\end{threeparttable}
\end{table*}

Table~\ref{tab:ablation_pool} concerns the sensitivity with Top-K options (K=3/7) and different graph regularizers, evaluated under both latent-ODE and \method. Overall, regularized graph construction consistently improves both metrics for the two frameworks, indicating that raw correlation graphs can be vulnerable to noise and volume conduction, while statistical regularization yields more reliable functional connectivity. Specifically, for latent-ODE, Graphical lasso and Norm regularization with K=3 achieve the strongest AUROC/Recall, suggesting that a sparser, regularized partial-correlation structure is preferable for continuous dynamics modeling. For \method, Norm with K=3 gives the best AUROC (0.881), whereas Graphical lasso with K=3 attains the highest Recall (0.613); the performance gap is small across K and regularizers, demonstrating robust behavior to graph-construction choices.

\begin{wraptable}{r}{0.5\linewidth}  
  \centering
  \caption{Ablation of GNN options on TUSZ (12s and 60s seizure detection) (AUROC↑, F1↑) Bold = best.}
  \label{tab:gnn_bc}
  \small
  \setlength{\tabcolsep}{2pt}
  \begin{tabular}{c l c c c}
    \toprule
    \multirow{2}{*}{ODE} & Method & T(Sec.) & AUROC & F1 \\
    \cmidrule(lr){2-5}
    \multirow{4}{*}{\rotatebox[origin=c]{90}{Temporal-spatial}}
      & EvolveGCN    & 12  &  0.791$\pm$0.003 &  0.401$\pm$0.002\\
      &           & 60 &  0.729$\pm$0.002 &  0.378$\pm$0.003\\
      \cmidrule(lr){2-5}
      & DCRNN    & 12  & 0.823$\pm$0.005 & 0.433$\pm$0.005\\
      &             & 60 & 0.818$\pm$0.004 & 0.417$\pm$0.007\\
      \cmidrule(lr){2-5}
     & GRU-GCN    & 12 & \bf 0.881$\pm$0.006 & \bf 0.496$\pm$0.017\\
      &             & 60 & \bf 0.828$\pm$0.003 & \bf 0.430$\pm$0.021\\
  \end{tabular}
\end{wraptable}

Table~\ref{tab:gnn_bc} shows the effects of GNN backbones on TUSZ under 12s and 60s forecasting horizons. We find that the GNN choice has a non-trivial impact on continuous seizure forecasting. GRU-GCN yields the best overall performance, reaching 0.881 AUROC / 0.496 F1 at 12s and 0.828 AUROC / 0.430 F1 at 60s. This indicates that recurrent gating over graph messages better captures fast and non-stationary ictal dynamics, especially for short-term prediction. DCRNN performs competitively but consistently below GRU-GCN (0.823/0.433 at 12s; 0.818/0.417 at 60s), suggesting diffusion-based spatiotemporal propagation is effective yet less expressive without explicit gating. In contrast, EvolveGCN degrades substantially, particularly for long-horizon forecasting (0.729 AUROC / 0.378 F1 at 60s), implying that merely evolving GCN parameters is insufficient under noisy epoch-wise correlation graphs. Overall, these results address Q4/W3 by demonstrating that ODEBRAIN’s continuous latent dynamics benefit most from temporally gated graph modeling, and the superiority is consistent across horizons.

\begin{wraptable}{r}{0.5\linewidth} 
  \centering
  \caption{Ablation of missing value (MV) on TUSZ (12s seizure detection) with AUROC↑, F1↑, and predicted missing graph structural similarity (Sim.)↑ (Bold = best).}
  \label{tab:missingvalue}
  \small
  \setlength{\tabcolsep}{2pt}
  \begin{tabular}{c l c c c}
    \toprule
    \multirow{2}{*}{MV} & Method & Sim. & AUROC & F1 \\
    \cmidrule(lr){2-5}
    \multirow{2}{*}{\rotatebox[origin=c]{90}{0\%}}
      & latent-ODE    & 0.53  &  0.791$\pm$0.003 &  0.401$\pm$0.002\\
      \cmidrule(lr){2-5}
     & \method    & \bf 0.63 & \bf 0.881$\pm$0.006 & \bf 0.496$\pm$0.017\\
    \midrule
    \multirow{2}{*}{\rotatebox[origin=c]{90}{30\%}}
      & latent-ODE    & 0.41  &  0.721$\pm$0.004 &  0.377$\pm$0.003\\
      \cmidrule(lr){2-5}
     & \method    & \bf 0.55 & \bf 0.845$\pm$0.002 & \bf 0.464$\pm$0.007\\
  \end{tabular}
\end{wraptable}

\begin{table*}[t]
\centering
\caption{Ablation on TUSZ dataset for 12s seizure detection with different top-$\tau$ options. \textbf{Bold} and \underline{underline} indicate best and second-best results.}
\label{tab:ab_top_tau}
\begin{tabular}{ccccc}  
\toprule
Top-$\tau$           & AUROC   & Recall      & F1  \\
\midrule
 2                 & 0.867 $\pm$ 0.003   & 0.575$\pm$0.003  & 0.484$\pm$0.009 \\
3                &  \textbf{0.881$\pm$0.006}  & \textbf{0.605$\pm$0.003}  & \textbf{0.496$\pm$0.017}  \\
7    & \underline{0.870$\pm$0.004}  & \underline{0.602$\pm$0.004}  & 0.488$\pm$0.013  \\
9          & 0.868$\pm$0.004  & 0.589$\pm$0.004  & 0.487$\pm$0.011  \\
11 & 0.866$\pm$0.004 & 0.571$\pm$0.002 & \underline{0.491$\pm$0.003} \\ 
13 & 0.865$\pm$0.003 & 0.562$\pm$0.004 & 0.474$\pm$0.003 \\

\bottomrule
\end{tabular}
\end{table*}

Table~\ref{tab:missingvalue} illustrates the robustness of \method when 30\% of EEG segments are randomly masked, comparing it with latent-ODE.
When 30\% segments are randomly masked, \method exhibits smaller AUROC drops from 0.881 to 0.845, and F1 from 0.496 to 0.464; exceeding the AUROC and F1 of latent-ODE by 0.124 and 0.067, respectively.
This demonstrates that \method maintains stable vector fields and detection performance under incomplete observations by leveraging adaptive gating operations within the vector field and stochastic regularization to suppress irregular time step jumps. The results indicate that \method achieves robustness to trajectory uncertainty under the effects of missing values, enhancing the capacity of ODE solvers.

Table~\ref{tab:ab_top_tau} shows the effects of the sparsity level of the correlation graph, controlled by the top-$\tau$ neighbors per node. Overall, AUROC remains stable performance across $\tau$ from 2 to 13 (0.865–0.881), indicating that \method is not overly sensitive to top-$\tau$ options. $\tau=3$ achieves the best AUROC (0.881) and F1 (0.496), while both too sparse ($\tau=2$) and too dense graphs ($\tau\ge 9$) lead to slight degradation. When the values of $\tau$ is small, the graph becomes too sparse making the edge GRU forward stage affect the quality of the graph descriptor. As $\tau$ increases, edges become much denser and correlation-based connectivity contains propagated noise, which makes the edge GRU forward more over-smoothing and injects noise structure into the initial state $\boldsymbol{z}_{0}$. The denser top-$\tau$ reduces the robustness of the vector field $f_{\theta}$. Therefore, we adopt $\tau=3$ as a good trade-off between predictive performance, robustness of the ODE dynamics.

\end{document}